\documentclass[10pt,twocolumn,letterpaper]{article}

\usepackage[pagenumbers]{cvpr}              %

\usepackage[accsupp]{axessibility} %
\usepackage[dvipsnames]{xcolor}
\usepackage{graphicx}
\usepackage{amsmath}
\usepackage{amssymb}
\usepackage{booktabs}
\usepackage{colortbl}
\usepackage{makecell}
\usepackage{multirow,multicol}
\usepackage{subcaption}
\usepackage{wrapfig,lipsum,booktabs}
\usepackage{bbm}
\usepackage{lipsum}
\usepackage{tabu}

\def\eg{\emph{e.g.}}
\def\ie{\emph{i.e.}}

\def\etal{\emph{et al.}}
\def\T{\mathsf{T}}

\newcolumntype{L}[1]{>{\raggedright\let\newline\\\arraybackslash\hspace{0pt}}m{#1}}
\newcolumntype{C}[1]{>{\centering\let\newline\\\arraybackslash\hspace{0pt}}m{#1}}
\newcolumntype{R}[1]{>{\raggedleft\let\newline\\\arraybackslash\hspace{0pt}}m{#1}}

\usepackage{amsmath,amsfonts,bm}

\def\eqref#1{equation~\ref{#1}}

\def\1{\bm{1}}

\def\va{{\bm{a}}}

\def\vk{{\bm{k}}}

\def\vq{{\bm{q}}}

\def\vu{{\bm{u}}}
\def\vv{{\bm{v}}}

\def\vx{{\bm{x}}}
\def\vy{{\bm{y}}}

\def\mA{{\bm{A}}}

\def\mH{{\bm{H}}}

\def\mK{{\bm{K}}}

\def\mU{{\bm{U}}}
\def\mV{{\bm{V}}}
\def\mW{{\bm{W}}}
\def\mX{{\bm{X}}}

\DeclareMathAlphabet{\mathsfit}{\encodingdefault}{\sfdefault}{m}{sl}
\SetMathAlphabet{\mathsfit}{bold}{\encodingdefault}{\sfdefault}{bx}{n}
\newcommand{\tens}[1]{\bm{\mathsfit{#1}}}

\def\tH{{\tens{H}}}

\def\tW{{\tens{W}}}

\definecolor{cvprblue}{rgb}{0.21,0.49,0.74}
\usepackage[pagebackref,breaklinks,colorlinks,citecolor=cvprblue]{hyperref}

\def\paperID{17028} %
\def\confName{CVPR}
\def\confYear{2024}

\title{Learning Correlation Structures for Vision Transformers}

\author{
Manjin Kim$^{1}$ \hspace{3mm} Paul Hongsuck Seo$^2$\thanks{Co-corresponding authors.} \hspace{3mm} Cordelia Schmid$^3$ \hspace{3mm} Minsu Cho$^1$\hspace{-0.4mm}\footnotemark[1]   \vspace{1.5mm}\\ 
$^1$POSTECH \hspace{18mm} $^2$Korea University \hspace{16mm} $^3$Google Research  \vspace{1.5mm}\\
{\tt \url{http://cvlab.postech.ac.kr/research/StructViT/}}
}

\begin{document}
\maketitle

\begin{abstract}

We introduce a new attention mechanism, dubbed structural self-attention (StructSA), that leverages rich correlation patterns naturally emerging in key-query interactions of attention. 
StructSA generates attention maps by recognizing space-time structures of key-query correlations via convolution and uses them to dynamically aggregate local contexts of value features. This effectively leverages rich structural patterns in images and videos such as scene layouts, object motion, and inter-object relations.
Using StructSA as a main building block, we develop the structural vision transformer (StructViT) and evaluate its effectiveness on both image and video classification tasks, achieving state-of-the-art results on ImageNet-1K, Kinetics-400, Something-Something V1 \& V2, Diving-48, and FineGym.
\end{abstract}

\section{Introduction}

How visual elements interact with each other in space and time is a crucial cue for visual understanding, \eg, recognizing actions in a video or analyzing scene layout patterns in an image. In computer vision, such relational patterns are effectively captured by the structure of correlations or similarities across visual elements in different positions~\cite{benabdelkader2004gait,selfsimilarity}; a correlation structure of an image reveals spatial layouts of similar  patterns~\cite{kang2021relational,kim2017fcss} and that of a video provides bi-directional motion likelihoods~\cite{rsa,selfy}. The ability to recognize those structural patterns may allow to better perform visual reasoning and generalize against challenging appearance variations and domain shifts~\cite{geirhos2021partial,tuli2021convolutional}.

\begin{figure}[t]
  \centering
  \includegraphics[width=\linewidth]{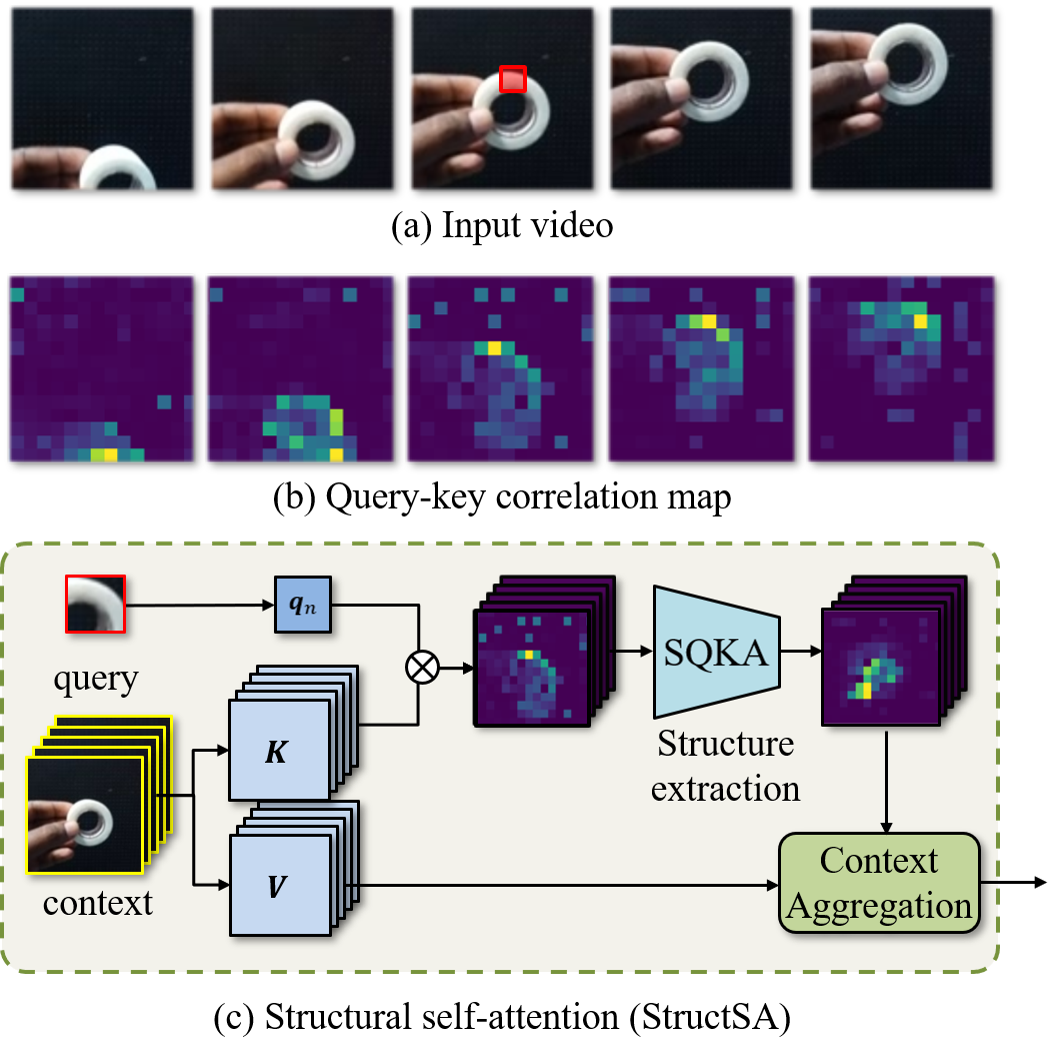}
  \vspace{-7mm}
   \caption{\textbf{Structural Self-Attention.} Given an input video and a query indicated by the red box in (a), the query-key correlation maps in (b) clearly reveal the structures of spatial layout and motion with respect to the query.
   The proposed attention mechanism in (c) is designed to leverage these rich structural patterns for computing attention scores in the self-attention process.
   }
   \label{fig:teasor}
  \vspace{-4mm}
\end{figure}

In this work, we introduce a novel self-attention mechanism, named \textit{structural self-attention} (StructSA), that effectively leverages diverse structural patterns for visual representation learning.
While the standard self-attention mechanism uses raw query-key correlations individually and ignores their geometric structures, %
the proposed StructSA recognizes diverse structural patterns from the correlations between the query and local chunks of keys via convolution and uses them to dynamically aggregate local contexts of value features, effectively capturing rich structural patterns such as scene layouts, object motion, and inter-object relations in images and video.
As illustrated in Fig.~\ref{fig:teasor} and detailed in Sec.~\ref{sec:method}, this is mainly achieved by empowering the standard self-attention mechanism with long-range convolutional interactions and dynamic contextual feature aggregation.  
To investigate the effect of StructSA, we also provide an in-depth analysis on the relationship to recent self-attention variants with convolutional projection~\cite{mvit,liang2021dualformer,mvitv2,pvt,cvt}, showing their potential and limitation in leveraging structural patterns. 

To validate the effect of StructSA, we develop the structural vision transformer that adopts it as a main building block, and perform an extensive set of experiments on both image and video classification tasks,
showing the effectiveness of learning structural patterns for visual representations. 
Our main contributions are as follows:
\begin{itemize}
\setlength\itemsep{0em}
    \item We introduce structural self-attention (StructSA) that learns correlation structures for visual representations with the vision transformer.
    
    \item We provide an in-depth analysis on the relationship between StructSA and self-attention variants with convolutional projections.
    
    \item Our new transformer network achieves state-of-the-art results on ImageNet-1K, Kinetics-400, Something-Something V1\&V2, Diving-48, and FineGym.
    
\end{itemize}

\section{Related Work}

\noindent\textbf{Transformer Networks in Vision.}
Since transformer networks~\cite{vaswani2017attention} showed remarkable success in natural language processing~\cite{brown2020language,devlin2018bert}, they have widely been adopted in various computer vision tasks as an alternative to CNNs~\cite{arnab2021vivit,carion2020end,vit,strudel2021segmenter,sun2019videobert}.
Despite of their success, the pure transformer networks require a large amount of training data compared to CNNs where convolution operations introduce desirable inductive biases such as locality and translation invariance allowing more efficient training~\cite{vit,raghu2021vision}.
This incentivized several methods to inherit the convolutional inductive biases via knowledge distillation~\cite{deit}, local self-attention~\cite{hu2019local,swin,ramachandran2019stand}, and architectural fusion~\cite{dai2021coatnet,cmt,uniformer,pvt,cvt}.
While recent methods using a convolutional projection~\cite{mvit,mvitv2,pvt,cvt} achieve remarkable improvements, we show that self-attention with a convolutional projection can be derived as a special form of our proposed method.

\smallskip
\noindent\textbf{Correlation Structure Modeling.}
Geometric structure of correlations between visual features, \ie, patterns of how they are similar to each other, allows us to understand relational patterns in visual data for various computer vision tasks.
Spatial self-correlation in images is used for suppressing photometric variations and revealing geometric layout of objects in the image~\cite{kang2021relational,kim2017fcss,selfsimilarity}.
Spatial cross-correlation between different images is often used for establishing semantic correspondences capturing structural similarities~\cite{han2017scnet,min2021convolutional,paul2018attentive}.
In the video domain, several methods exploit the structure of spatial cross-correlations between consecutive frames to estimate optical flow~\cite{dosovitskiy2015flownet,yang2019volumetric} or to learn motion features for action recognition~\cite{motionsqueeze,corrnet}.
Kwon \etal~\cite{selfy} propose spatio-temporal self-correlations for learning bi-directional motion features and Kim \etal~\cite{rsa} introduce relational self-attention that generates attention weights dynamically from the structure of the spatio-temporal self-correlations.
However, these two methods use self-correlations between the query and its local spatio-temporal neighborhoods only, thus, are limited in learning global relational patterns between distant features.
Inspired by this, we introduce structural self-attention that capturing not only the spatio-temporal local self-correlation but also cross-correlations between features in the distance, utilizing both motion and global spatio-temporal inter-feature relations for learning motion-centric video representations.

\section{Our Approach}~\label{sec:method}
We propose a novel self-attention mechanism, named \textit{structural self-attention} (StructSA), that is designed to leverage rich correlation structures naturally emerging in key-query interactions of attention. 
We start by revisiting the vanilla self-attention and its limitation and then describe the details of StructSA. We also provide an in-depth analysis of recent self-attention variants with convolutional projections from the perspective of learning structural patterns.

\subsection{Background: Self-Attention}

Self-attention (SA)~\cite{vaswani2017attention} is a primitive operation for modern transformer networks~\cite{arnab2021vivit,vit,deit}.
Given $N$ input features $\bm{X}=[\bm{x}_1,\cdots,\bm{x}_N]\in \mathbb{R}^{N\times C}$, SA first projects the input $\bm X$ linearly into queries, keys, and values, and transforms each $C$-dimensional input feature $\vx_i$ into a contextualized output feature $\vy_i$ by
\vspace{-2mm}
\begin{gather}
    \vy_i = 
    \sigma \left( \vq_i {\mK}^\mathsf{T} \right) \mV = \sum^N_j \sigma_j \left( \vq_i {\vk_j}^\mathsf{T} \right) \vv_j  \in \mathbb{R}^{1 \times C}, \label{eq:self_attention} \\ %
    \vq_i = \vx_i \mW^{\mathrm{Q}},~~~~
    \mK = \mX \mW^{\mathrm{K}} , ~~~~
    \mV = \mX \mW^{\mathrm{V}} \nonumber
\end{gather}
where $\sigma$ is a softmax function and  $\mW^Q, \mW^K, \mW^V\in \mathbb{R}^{C\times C}$ are projection matrices for query $\vq_i\in \mathbb{R}^{1 \times C}$, keys $\mK \in \mathbb{R}^{N \times C}$, and values $\mV \in \mathbb{R}^{N \times C}$, respectively.
Here we use a 1-dimensional sequence of input features for notational simplicity, and the operation can be extended to a larger dimensionality.
After computing a correlation map $\vq_i {\mK}^\mathsf{T}$, the vanilla SA uses individual correlation values independently, \ie, $\vq_i {\vk_j}^\mathsf{T}$, for value aggregation while ignoring the \textit{structure} of the map, which leads to the same output regardless of the order of features.
This permutation invariance prevents SA from capturing spatial layouts~\cite{hu2019local,selfsimilarity} or motions~\cite{rsa,selfy,liu2019learning} of objects in images or videos.
Positional encoding for SA helps spatial awareness, but 
the {\em structures} of correlations are still not recognized in value aggregation~\cite{cordonnier2019relationship,dai2019transformer,rsa}.

\subsection{Structural Self-Attention}
\label{sec:structsa}
We introduce a novel self-attention mechanism, named \textit{structural self-attention} (StructSA), that effectively incorporates rich structural patterns of query-key correlation into  contextual feature aggregation.
The StructSA mechanism consists of two steps: (i) structural query-key attention and (ii) contextual value aggregation. 
Unlike the vanilla query-key attention where individual correlation values themselves are used as attention scores, the {\em structural query-key attention} takes the correlation map as a whole and detect structural patterns from it in attention scoring. The subsequent {\em contextual value aggregation} then combines the attention scores together to compute diverse sets of kernel weights that are used for dynamically collecting local contexts of value features.

\smallskip
\noindent\textbf{Structural Query-Key Attention.}
To transform the vanilla query-key attention into structure-aware one, the structural query-key attention (SQKA) deploy convolutions on top of query-key correlation $\vq_i \mK^\T$:
\begin{align}
    \mA_i = \sigma \left( \texttt{conv} \left( \vq_i \mK, \mU^{\mathrm{K}} \right) \right) \in \mathbb{R}^{N \times D},
\end{align}
where $\mU^{\mathrm{K}} \in \mathbb{R}^{M \times D}$ is $D$ convolutional kernels with size $M$.
Note that $\sigma$ is a softmax function taken over all $ND$ entries in the input matrix; we observe that it is empirically more stable compared to $D$ individual softmax operations over $N$ entries.
Each element of $\mA_i$ is computed as
\begin{align}
\va_{i,j} &= \sigma_{j} \left( \vq_i \mK_j^\mathsf{T} {\mU^{\mathrm{K}}} \right) \in \mathbb{R}^{1 \times D}, \\
    \mK_j&=\mX_{(j)} \mW^{\mathrm{K}} \in  \mathbb{R}^{M \times C}, \nonumber %
\end{align}
where $\sigma_{j}$ returns a $D$-dimensional softmax-ed output for $j$th location and $\mX_{(j)} \in \mathbb{R}^{M\times C}$ is local context features whose context window is centered at $j$.

Unlike the vanilla query-key attention, which is agnostic to its neighborhood structure, SQKA is empowered by convolution to recognize a local correlation structure of $\vq_i \mK_j^\T \in \mathbb{R}^{1\times M}$ and  transform it into a $D$-dimensional vector; the convolution kernels $\mU^\mathrm{K}$ act as \textit{correlation pattern detectors}.
In particular, when $i=j$, the correlation map reduces to local self-similarity~\cite{selfsimilarity} that is known to be effective for capturing spatial layout patterns~\cite{selfsimilarity} or spatio-temporal motion~\cite{rsa,selfy}, meaning that SQKA recognizes diverse correlation patterns including  self-similarity~\cite{selfsimilarity} via long-range interaction between query $i$ and context $j$.

\smallskip
\noindent\textbf{Contextual Value Aggregation.}
Given SQKA recognizing structural patterns of correlation, 
StructSA combines SQKA entries into a weight $\kappa_{i,j}^{\mathrm{struct}}$ to aggregate value $\vv_j$: 
\begin{align}
    \vy_i = \sum_{j=1}^N \sigma_{j} \left( \vq_i \mK_j^\mathsf{T} {\mU^{\mathrm{K}}} \right) {\vu^{\mathrm{V}}}^\mathsf{T} \vv_j = \sum_{j=1}^N \kappa_{i,j}^{\mathrm{struct}} \vv_j, \label{eq:structsa}
\end{align}
where $\vu^{\mathrm{V}} \in \mathbb{R}^{1 \times D}$ is a vector that linearly combines $D$ pattern scores to the final attention weight.

We further extend Eq.~(\ref{eq:structsa}) to generate a spatial kernel not a single scalar for each position $j$.
We call this method contextual aggregation and it has a following form:
\begin{gather}
    \vy_i = \sum_{j=1}^N \sigma_{j} \left( \vq_i \mK_j^\mathsf{T} \mU^{\mathrm{K}} \right) {\mU^{\mathrm{V}}}^\T \mV_j = \sum_{j=1}^N \bm \kappa_{i,j}^{\mathrm{struct}} \mV_j,\\
    \mV_j = \mX_{(j)} \mW^{\mathrm{V}} \in  \mathbb{R}^{M \times C}. \nonumber
\end{gather}
Compared to $\vu^{\mathrm{V}}$ that produces a scalar weighting a single element $\vv_j$, $\mU^{\mathrm{V}} \in \mathbb{R}^{M \times D}$ generates spatial kernel dynamically aggregating a local context of value $\mV_j$ for every position $j$;
each column of $\mU^\mathrm{V}$, \ie, $\mU^\mathrm{V}_{:,d} \in \mathbb{R}^{M \times 1}$, plays a role as a {\em context aggregator} that performs a weighted pooling of local context $\mV_j$ and thus different combinations of these context aggregators result in diverse dynamic kernels $\bm \kappa_{i,j}^{\mathrm{struct}}$ for different locations $j$.

\begin{figure*}[t]
  \begin{subfigure}{1\textwidth} %
    \refstepcounter{subfigure}\label{fig:in1k_qual:a}
    \refstepcounter{subfigure}\label{fig:in1k_qual:b}
    \refstepcounter{subfigure}\label{fig:in1k_qual:c}
    \end{subfigure}
  \centering
  \includegraphics[width=0.85\linewidth]{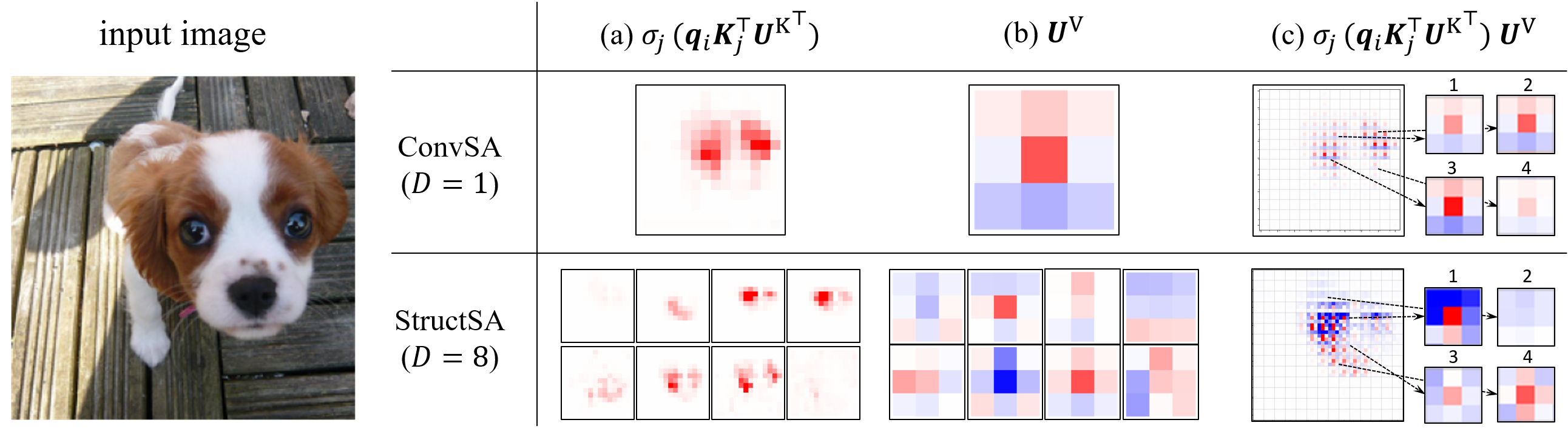}
  \vspace{-2mm}
  \caption{\textbf{Visualization of ConvSA and StructSA on ImageNet-1K.}
  The query location $i$ is set to the center of the image and the kernel size $M=3\times3$. 
  Given the left input image, we compare ConvSA ($D=1$) and StructSA ($D=8$) in terms of (a) $D$ attention maps $\sigma_{jD} ( \vq_i \mK_j^\mathsf{T} {\mU^{\mathrm{K}}}^\mathsf{T})$, (b) local feature aggregation patterns learned in $\mU^V$, and (c) the combinations of (a) and (b).
  Note that in (c), each location $j$ has an aggregation map of the kernel size $M=3\times3$ and thus we also show enlarged images for four different sample locations~$j$.
  }
  \label{fig:in1k_qual}
  \vspace{-2mm}
\end{figure*}

\subsection{Relationship to Convolutional Self-Attention}
\label{sec:structure_convsa}
Recent vision transformers~\cite{mvit,cmt,mvitv2,pvt,cvt} often adopt convolutional inductive biases in the form of self-attention with convolutional projections (ConvSA).
In this section, we analyze ConvSA with a lens of StructSA and show its potential and limitation for learning structures from query-key correlations.
Different from SA, ConvSA computes project keys and values $\mK^\mathrm{conv},\mV^\mathrm{conv}  \in \mathbb{R}^{N \times C}$
using a convolution operation over the input feature map $\mX$:
\begin{align}
    \mK^{\mathrm{conv}} &= \left[ \vk_1^{\mathrm{conv}}, \cdots, \vk_N^{\mathrm{conv}} \right] = \texttt{conv}(\mX, \tW^{\mathrm{K}}),\\
    \mV^{\mathrm{conv}} &= \left[ \vv_1^{\mathrm{conv}}, \cdots, \vv_N^{\mathrm{conv}} \right] = \texttt{conv}(\mX, \tW^{\mathrm{V}}),
\end{align}
where $\texttt{conv}$ is a convolution operation, and $\tW^{\mathrm{K}},\tW^{\mathrm{V}} \in \mathbb{R}^{M \times C \times C}$ are kernel weights with a kernel size $M$ for key and value projections, respectively. %

In most previous methods, ConvSA is implemented with a channel-wise separable convolution~\cite{howard2017mobilenets}, which consists of two factorized convolution operations, \ie, point-wise and channel-wise convolutions~\cite{mvit,cmt,mvitv2,pvt,cvt}.
In this case, each key $\vk^{\mathrm{conv}}_i$ and value $\vv^{\mathrm{conv}}_i$ is computed from a local context $\mX_i$ by
\begin{align}
    \vk_i^{\mathrm{conv}} &= {\vu^{\mathrm{K}}}^\T \mX_i \mW^{\mathrm{K}} = {\vu^{\mathrm{K}}}^\T \mK_i  \in \mathbb{R}^{1 \times C}, \label{eq:conv_proj_key}\\
    \vv_i^{\mathrm{conv}} &= {\vu^{\mathrm{V}}}^\T \mX_i \mW^{\mathrm{V}} = {\vu^{\mathrm{V}}}^\T \mV_i  \in \mathbb{R}^{1 \times C}, \label{eq:conv_proj_value}
\end{align}
where $\mW^{\mathrm{K}}, \mW^{\mathrm{V}} \in \mathbb{R}^{C \times C}$ are weights for the linear projection that are equivalent to point-wise convolution, and $\vu^{\mathrm{K}}, \vu^{\mathrm{V}} \in \mathbb{R}^{M \times 1}$ are channel-wise convolution weights that are used to spatially aggregate linearly projected context $\mK_i$ and $\mV_i$, respectively.
Note that here we assume the channel-wise convolution weights are shared across channels for simplicity without loss of generality and the full derivation is available in Appendix~\ref{sec:supp_full_derivation}.

From Eq.~(\ref{eq:self_attention}) combined with Eq.~(\ref{eq:conv_proj_key}) and (\ref{eq:conv_proj_value}), a transformed output $\vy_i$ in ConvSA is obtained by
\begin{align}
    \vy_i &= 
    \sum_{j=1}^N \sigma_j \left( \vq_i {\vk_j^{\mathrm{conv}}}^\mathsf{T}  \right) \vv_j^{\mathrm{conv}} \nonumber \\
    &= \sum_{j=1}^N \sigma_j \left( \vq_i \mK_j^\mathsf{T} \vu^{\mathrm{K}} \right) {\vu^{\mathrm{V}}}^\T \mV_j = \sum_{j=1}^N \bm \kappa_{i,j}^{\mathrm{conv}} \mV_j ,
    \label{eq:convolutional_self_attention}
\end{align}
where $\sigma_j$ is $j$th entry of the softmax over $N$ tokens.
This reveals that an attention score $\sigma_j ( \vq_i {\vk_j^{\mathrm{conv}}}^\mathsf{T} )$
is computed by projecting a local correlation map $\vq_i \mK_j^\T \in \mathbb{R}^{1\times M}$ by $\vu^\mathrm{K}$, and a dynamic kernel $\bm \kappa_{i,j}^{\mathrm{conv}}$ for the final feature aggregation of $\mV_j$ is obtained by weighting the aggregation pattern presented in $\vu^\mathrm{V}$ using the computed attention map.
Given that correlation map $\vq_i \mK_j^\T$ represents a structural pattern, we can interpret that $\vu^\mathrm{K}$ acts as a {\em pattern detector} that extracts a specific correlation pattern from $\vq_i \mK_j^\T$, whereas $\vu^\mathrm{V}$ plays a role as a {\em context aggregator} that performs a weighted pooling of local context $\mV_j$.
Due to the presence of this pattern detector $\vu^\mathrm{K}$ and its corresponding context aggregator $\vu^\mathrm{V}$, ConvSA can leverage a structural pattern of input for context aggregation.

\smallskip
\noindent\textbf{Limitation.}
Although ConvSA can learn, unlike SA, a structural pattern over correlation maps by $\vu^\mathrm{K}$, it only learns a single pattern and encodes various shapes in correlation maps into a scalar value representing the similarity against the learned pattern; as the result, the final dynamic kernel $\bm \kappa_{i,j}^\mathrm{conv}$ for every $j$ reduces to the identical pattern of  $\vu^\mathrm{V}$ with different weighting only.
This lack of expressiveness in $\vu^\mathrm{K}$ and $\vu^\mathrm{V}$ prevents ConvSA from capturing diverse structural patterns and generating diverse dynamic kernels.
In constrast, StructSA learns $D$ different pattern extractors in $\mU^\mathrm{K}$ and represents various local correlation shapes by a set of $D$ similarity scores.
These scores are then combined with the $D$ context aggregators in $\mU^\mathrm{V}$; %
different combinations of these context aggregators result in diverse dynamic kernels $\bm \kappa_{i,j}^{\mathrm{struct}}$ for different locations $j$.

\smallskip
\noindent\textbf{Visualization.} The aforementioned difference between ConvSA and StructSA, as well as their effects, can be better understood by visualizing the kernel computation process. Figure~\ref{fig:in1k_qual} provides such a visual comparison of how structural patterns are used in ConvSA and StructSA given an example image from ImageNet-1K~\cite{deng2009imagenet}.
From a query-key correlation map, ConvSA generates a single attention map (Fig.~\ref{fig:in1k_qual:a}, top). 
These scores are then combined with the context aggregator $\vu^\mathrm{V}$ (Fig.~\ref{fig:in1k_qual:b}, top), which conveys only a single aggregation pattern. This causes local features to be aggregated with the identical pattern in $\vu^\mathrm{V}$ for all locations, and the only difference remains in their scales (Fig.~\ref{fig:in1k_qual:c}, top).
In contrast, StructSA generates diverse attention maps using $D$ pattern detectors, each capturing different structures in the query-key correlation maps (Fig.~\ref{fig:in1k_qual:a}, bottom), and combines them with different context aggregators (Fig.~\ref{fig:in1k_qual:b}, bottom) resulting in rich aggregation patterns for different locations $j$ (Fig.~\ref{fig:in1k_qual:c}, bottom).
For more in-depth comparison, please refer to Appendix~\ref{sec:supp_visualization}.

\section{Experiments}
To validate the effectiveness of the proposed method on visual representation learning,
we conduct extensive experiments on image and video classification benchmarks.

\subsection{Experimental Setup}
\paragraph{Datasets.} \textbf{ImageNet-1K}~\cite{deng2009imagenet} is a large-scale dataset with 1.2M images labeled by 1000 object classes.  
\textbf{Kinetics-400}~\cite{kay2017kinetics} is one of the most popular large-scale video datasets with 400 action classes. We use 241k action clips available online.
\textbf{Something-Something-V1 \& V2}~\cite{goyal2017something,mahdisoltani2018effectiveness} are both large-scale action recognition benchmarks, including 108k and 220k action clips, respectively. Both datasets share the same motion-centric action classes, e.g., ‘pushing something from left to right,’ so thus capturing fine-grained motion is crucial to achieving the better performance.
\textbf{Diving-48}~\cite{li2018resound} is a fine-grained action benchmark that is heavily dependent on temporal modeling [3], containing 18k videos with 48 diving classes.
\textbf{FineGym}~\cite{shao2020finegym} is a motion-centric benchmark that includes gymnastics action classes with severe deformations.

\label{sec:experimental_setup}
\paragraph{Training \& Testing Protocols.}
For image classification, we follow the training strategy of  DeiT~\cite{deit} adopting random clipping, random horizontal flipping, mixup~\cite{zhang2017mixup}, cutmix~\cite{yun2019cutmix}, random erasing~\cite{zhong2020random} and label-smoothing~\cite{muller2019does} to augment the input images for training.
We train all models from scratch for 300 epochs using AdamW optimizer~\cite{loshchilov2017decoupled} with a cosine learning rate schedule including 5 warm-up epochs.
The batch size, learning rate, and weight decay are set to 1024, 1e-3, and 0.05, respectively.
For comparison on stronger experiment setup~\cite{lvvit,uniformer,mvitv2,yuan2022volo}, we also train our models using Token Labeling~\cite{lvvit} and larger resolution images, \ie, $384\times384$, following the protocols in \cite{uniformer}.

For video classification, we follow training protocols and data augmentation recipes in MViT~\cite{mvit}.
For Kinetics-400, we sample 16 or 32 frames using the dense sampling strategy~\cite{wang2018non}. 
We temporally inflate the model weights pretrained on ImageNet-1K and finetune it for 110 epochs including 10 warm-up epochs.
We use AdamW~\cite{loshchilov2017decoupled} optimizer with the cosine learning rate schedule.
We set the batch size, learning rate, weight decay, and stochastic depth rate to 64, 2e-4, 0.05, and 0.1, respectively.
For Something-Something, Diving-48, and FineGym, we utilize the segment-based sampling strategy~\cite{tsn} and do not use the random horizontal flip for data augmentation.
We initialize the model with the weights pretrained on Kinetics-400 and finetune the model for 60 epochs including 5 warm-up epochs.
Other training hyperparameters are the same as those for Kinetics-400.
For testing, we sample multiple clips at different temporal indices for each clip or cropping different spatial regions and then obtain the final score by computing an average over the scores for each clip.
We train all models once using 8 to 16 NVIDIA A100 GPUs. %

\smallskip
\noindent{\textbf{Metrics.}}
We measure top-1 and top-5 accuracy as performance metrics, except for FineGym, we compute averaged per-class top-1 accuracy. As efficiency metrics, we measure the number of parameters and FLOPs.

\subsection{Analysis of StructSA}

\begin{table}[t]
\begin{subtable}[t]{\columnwidth}
    \centering  
    \captionsetup{width=\columnwidth}
    \scalebox{0.9}{
    \begin{tabular}[t]{C{2.5cm}|C{1.1cm}C{1.1cm}|C{1.1cm}C{1.1cm}}
    \Xhline{1.0pt}
    \multirow{2}{*}{$D$} & \multicolumn{2}{c|}{ImageNet-1K} & \multicolumn{2}{c}{Something V1} \\
    &  top-1 & top-5 & top-1 & top-5 \\
    \Xhline{0.7pt}
    0 & 80.5 & 95.0 & 48.3 & 76.6 \\
    \Xhline{0.3pt}
    1 & 80.8 & 95.2 & 49.7 & 77.6 \\
    2 & 80.9 & 95.2 & 50.1 & 78.0 \\
    4 & 81.1 & 95.4 & 50.4 & 78.2 \\
    8 & \textbf{81.3} & \textbf{95.5} & \textbf{50.6} & \textbf{78.5} \\
    \Xhline{1.0pt}
    \end{tabular}
    }
    \vspace{-0.5mm}
    \caption{\textbf{Structure dimension $D$}. 
    }
    \label{tab:ablation_d}
\end{subtable}

\vspace{1.5mm}

\begin{subtable}[t]{\columnwidth}
    \centering  
    \captionsetup{width=\columnwidth}
    \scalebox{0.9}{
    \begin{tabular}[t]{C{2.5cm}|C{1.1cm}C{1.1cm}|C{1.1cm}C{1.1cm}}
    \Xhline{1.0pt}
    \multirow{2}{*}{$M$} & \multicolumn{2}{c|}{ImageNet-1K} & \multicolumn{2}{c}{Something V1} \\
    & top-1 & top-5 & top-1 & top-5 \\
    \Xhline{0.7pt}
    - & 80.5 & 95.0 & 48.3 & 76.6 \\
    \Xhline{0.3pt}
    $1 \times 1 ~(\times 1)$ & 80.6 & 95.0 & 48.5 & 76.9 \\
    $3 \times 3 ~(\times 3)$ & \textbf{81.1} & \textbf{95.4} & 50.4 & \textbf{78.2} \\
    $5 \times 5 ~(\times 5)$ & \textbf{81.1} & \textbf{95.4} & \textbf{50.5} & \textbf{78.2} \\
    $7 \times 7 ~
    (\times 7)$ & 81.0 & 95.2 & \textbf{50.5} & 78.1 \\
    \Xhline{1.0pt}
    \end{tabular}
    }
    \vspace{-0.5mm}
    \caption{\textbf{Kernel size $M$.} 
    }
    \label{tab:ablation_m_s}
\end{subtable}

\vspace{1.5mm}

\begin{subtable}[t]{\columnwidth}
    \centering  
    \captionsetup{width=\columnwidth}
    \scalebox{0.9}{
    \begin{tabular}[t]{C{2.5cm}|C{1.1cm}C{1.1cm}|C{1.1cm}C{1.1cm}}
    \Xhline{1.0pt}
    \multirow{2}{*}{aggregation} & \multicolumn{2}{c|}{ImageNet-1K} & \multicolumn{2}{c}{Something V1} \\
    & top-1 & top-5 & top-1 & top-5 \\
    \Xhline{0.7pt}
    - & 80.5 & 95.0 & 48.3 & 76.6 \\
    \Xhline{0.3pt}
    element & 80.9 & 95.2 & 49.6 & 77.5 \\
    context & \textbf{81.1} & \textbf{95.4} & \textbf{50.4} & \textbf{78.2} \\
    \Xhline{1.0pt}
    \end{tabular}
    }
    \vspace{-0.5mm}
    \caption{\textbf{Context aggregation method.} 
    }
    \label{tab:ablation_context_agg}
\end{subtable}

\vspace{-1mm}
\caption{\textbf{Ablation studies on ImageNet-1K and Something-Something V1}. Top-1 and top-5 accuracies (\%) are shown.
Otherwise specified, we use 16 frames as input and set $D=4$, $M=3\times3\times3$, and patch-wise context aggregation as default.
}
\vspace{-2mm}
\end{table}

StructSA can be readily integrated into any existing ViTs to enhance visual representations by capturing correlation structures.
In this subsection, we experimentally validate and analyze the impact of StructSA. %
Here, for a direct comparison with SA, we choose to use DeiT-S~\cite{deit} as the baseline backbone; DeiT is a pure SA-based vision transformer and thus adequate for validating the effect of StructSA, avoiding any intervention of additional components.
In this analysis, we replace all the SA layers in DeiT with StructSA layers.
The evaluations are done on ImageNet-1K~\cite{deng2009imagenet} and Something-Something-V1~\cite{goyal2017something} benchmarks while varying the structure dimension $D$, the kernel size $M$, and context aggregation methods.
We follow the training and testing protocols in Sec.~\ref{sec:experimental_setup}, except that we directly finetune the ImageNet-1K-pretrained model on Something-Something-V1 using random cropping only for data augmentation.

\smallskip
\noindent\textbf{Structure Dimension $D$.} Table~\ref{tab:ablation_d} shows the effect of the structure dimension $D$.
Compared to the baseline with the vanilla SA ($D=0$), applying ConvSA ($D=1$) improves the performance as shown in \cite{cpvt,cvt}.
As we increase $D$ from 1 to 8, we obtain gradual improvements up to 0.5\%p and 0.9\%p at top-1 accuracy on ImageNet-1K and Something-Something-V1.
This confirms the limitation of ConvSA and the effectiveness of StructSA.

\smallskip
\noindent\textbf{Kernel Size $M$.} In Table~\ref{tab:ablation_m_s}, we also investigate different kernel sizes $M$. %
Compared to the baseline, the model with the kernel size $M=1\times1\times1$ performs similar accuracies on both datasets whereas that with $M=3\times3\times3$ improves the performance dramatically; it validates the effectiveness of learning geometric structures.
The performance saturates as the kernel size gets larger than $5\times5\times5$.

\smallskip
\noindent\textbf{Context Aggregation Method.} In Table~\ref{tab:ablation_context_agg}, we also compare different context aggregation method. As a result, patch-wise aggregation performs 0.2\%p and 0.7\%p at top-1 accuracy on ImageNet-1K and Something-Something-V1.
For more ablation experiments, please refer to our Appendix~\ref{sec:supp_ablation}.

\subsection{Comparison to State of the Art}

\subsubsection{Structural Vision Transformer (StructViT)}
To build an advanced vision transformer considering the recent development of multiscale representation learning~\cite{mvit,uniformer,mvitv2,pvt,t2tvit}, we integrate StructSA into UniFormer~\cite{uniformer}. 
The transformer network, dubbed the Structural Vision Transformer (StructViT) is constructed by replacing all vanilla self-attention in UniFormer with StructSA as a main building block.
It takes as input either a video clip or an image $\Tilde{\bm{X}} \in \mathbb{R}^{T \times H \times W \times 3}$, where $T$, $H$, and $W$ denotes the temporal length, height, and width of the input, respectively.
For images as input, we set the temporal length $T=1$.
Before being fed into our networks, the input video is tokenized into overlapping 3D tublets of size $3 \times 4 \times 4 \times 3$ with a stride of $2 \times 4 \times 4$ while the input image is into non-overlapping 2D patches of size $4 \times 4 \times 3$.

\begin{table}[t]
\vspace{-4mm}
\centering
\captionsetup{width=\columnwidth}
\setlength\tabcolsep{1.3pt}
\scalebox{0.9}{
\begin{tabular}[t]{ccccc}
\Xhline{1.0pt}
model & type & \# blocks & \# channels (\# heads) \\
\Xhline{0.7pt}
StructViT-S & [C,C,S,S] & [3, 4, 8, 3] & [64, 128, 320 (5), 512 (8)]  \\
StructViT-B & [C,C,S,S] & [5, 8, 20, 7] & [64, 128, 320 (5), 512 (8)] \\
StructViT-L & [C,C,S,S] & [5, 10, 24, 7] & [128, 192, 448 (7), 640 (10)] \\
\Xhline{1.0pt}
\end{tabular}
}
\vspace{-3mm}
\caption{\textbf{Configurations of StructViT variants}. "C" and "S" denote a convolution and StructSA block, respectively.}
\label{tab:architecture}
\vspace{-3mm}
\end{table}

Our networks comprise four stages, each of which has multiple neural blocks, and leverage a hierarchical design with decreasing resolutions and increasing number of channels from early to late stages following \cite{uniformer}.
For the first two stages, each block consists of a conditional positional encoding layer~\cite{cpvt}, a convolutional layer, and an MLP, whereas the convolutional layer is replaced by a StructSA layer in the blocks in the last two stages.
Note that when we employ StructSA, we use multi-head configurations and do not share weights across channels for channel-wise convolutions.
We build three different StructViT architectures as shown in Table~\ref{tab:architecture}. %
Our models are comparable to UniFormers~\cite{uniformer} in the same sizes as our model configurations are based on UniFormer's; adding the structure dimension $D$ in StructSA introduces only few additional parameters.

In practice, StructSA introduces additional FLOPs for processing instances compared to the vanilla SA.
One way of building an efficient StructSA is to adopt a larger stride in the key/value projections, which effectively reduces the number of keys and values~\cite{mvit,mvitv2}.
We test a few variants with a larger stride to see the performances of StructViT with matching FLOPs with their corresponding UniFormer architectures.
We denote each model with StructViT-$X$-$D$-$S$ where $X$, $D$, and $S$ represent the architecture size, the structure dimension, and the stride, respectively.
For training, we use stochastic depth~\cite{huang2016deep} with the probability of 0.1/0.3/0.4 for StructViT-S/B/L, respectively.
We use random cropping only for StructViT-B on Something-Something, Diving-48, and FineGym.
We use 8 NVIDIA A100 GPUs for training StructViT-S/B and 16 GPUs for StructViT-L.
We follow the protocols in Sec.~\ref{sec:experimental_setup} for the rest.

\begin{table}[t]
	\vspace{-4mm}
\centering
    \scalebox{0.9}{
    	\begin{tabular}{L{4cm}|C{1.2cm}|C{1.1cm}|C{1.1cm}}
    	    \Xhline{1.0pt}
    		\multirow{2}*{method} & \#params & FLOPs & IN1K \\
    		 ~ & (M) & (G) & top-1  \\
    	    \Xhline{0.7pt}
    		EffcientNet-B5~\cite{tan2019efficientnet} & 30 & 9.9 & 83.6 \\
                ConvNext-T~\cite{convnext} & 29 & 4.5 & 83.1 \\
    		\Xhline{0.3pt}
    		DeiT-S~\cite{deit} & 22 & 4.6 & 79.9 \\
    		PVT-S~\cite{pvt} & 25 & 3.8 & 79.8 \\
    		Swin-T~\cite{swin} & 29 & 4.5 & 81.3 \\
    		Focal-T~\cite{focal} & 29 & 4.9 & 82.2 \\
    		CSwin-T~\cite{cswin} & 23 & 4.3 & 82.7 \\
    		\Xhline{0.3pt}
    		CvT-13~\cite{cvt} &  20 & 4.5 & 81.6 \\
    		CoAtNet-0~\cite{dai2021coatnet} &  25 & 4.2 & 81.6 \\
    		LV-ViT-S \cite{lvvit} & 26 & 6.6 & 83.3 \\
    		UniFormer-S~\cite{uniformer} & 22 & 3.6 & 82.9 \\
    		UniFormer-S* $\uparrow$384~\cite{uniformer} & 22 & 11.9 & 84.6 \\
    		MViTv2-S~\cite{mvitv2} & 24 & 4.7 & 82.3\\
    		\cellcolor{gray!20}{StructViT-S-4-2} &  \cellcolor{gray!20}{23} & \cellcolor{gray!20}{3.6} & \cellcolor{gray!20}{82.9} \\
    		\cellcolor{gray!20}{StructViT-S-4-1} &  \cellcolor{gray!20}{23} & \cellcolor{gray!20}{4.3} & \cellcolor{gray!20}{83.2} \\
    		\cellcolor{gray!20}{StructViT-S-8-1} &  \cellcolor{gray!20}{24} & \cellcolor{gray!20}{5.4} & \cellcolor{gray!20}{83.3} \\
            \cellcolor{gray!20}{StructViT-S-4-1*} &  \cellcolor{gray!20}{23} & \cellcolor{gray!20}{4.3} & \cellcolor{gray!20}{84.0} \\
            \cellcolor{gray!20}{StructViT-S-4-1* $\uparrow$384} &  \cellcolor{gray!20}{23} & \cellcolor{gray!20}{17.3} & \cellcolor{gray!20}{\textbf{85.2}} \\
    	    \Xhline{1.0pt}
    		EffcientNet-B7 \cite{tan2019efficientnet} & 66 & 39.2 & 84.3 \\
                ConvNext-B~\cite{convnext} & 89 & 15.4 & 83.8 \\
                ConvNext-B $\uparrow$384~\cite{convnext} & 89 & 45.0 & 85.1 \\
    		\Xhline{0.3pt}
    		PVT-L~\cite{pvt} & 61 & 9.8 & 81.7 \\
    		Swin-S~\cite{swin} & 50 & 8.7 & 83.0 \\
    		Focal-S~\cite{focal} & 51 & 9.1 & 83.5 \\
    		CSwin-S~\cite{cswin} & 35 & 6.9 & 83.6 \\
    		\Xhline{0.3pt}
    		CvT-21~\cite{cvt} &  32 & 7.1 & 82.5 \\
    		CoAtNet-1~\cite{dai2021coatnet} &  42 & 8.4 & 83.3 \\
    		LV-ViT-M \cite{lvvit} & 56 & 16.0 & 84.1 \\
            UniFormer-B~\cite{uniformer} & 50 & 8.3 & 83.8 \\
            UniFormer-B* $\uparrow$384~\cite{uniformer} & 50 & 27.2 & 86.0 \\
            MViTv2-S~\cite{mvitv2} & 35 & 7.0 & 83.6\\
            MViTv2-B~\cite{mvitv2} & 52 & 10.2 & 84.4\\
            MViTv2-B $\uparrow$384~\cite{mvitv2} & 52 & 36.7 & 85.2\\
    		\cellcolor{gray!20}{StructViT-B-4-2} &  \cellcolor{gray!20}{51} & \cellcolor{gray!20}{8.3} & \cellcolor{gray!20}{84.0} \\
    		\cellcolor{gray!20}{StructViT-B-4-1} &  \cellcolor{gray!20}{51} & \cellcolor{gray!20}{9.9} & \cellcolor{gray!20}{84.2} \\
    		\cellcolor{gray!20}{StructViT-B-8-1} &  \cellcolor{gray!20}{52} & \cellcolor{gray!20}{12.0} & \cellcolor{gray!20}{84.3} \\
            \cellcolor{gray!20}{StructViT-B-4-1*} &  \cellcolor{gray!20}{51} & \cellcolor{gray!20}{9.9} & \cellcolor{gray!20}{85.4} \\
      \cellcolor{gray!20}{StructViT-B-4-1* $\uparrow$384} &  \cellcolor{gray!20}{51} & \cellcolor{gray!20}{40.7} & \cellcolor{gray!20}{\textbf{86.5}} \\
    		\Xhline{1.0pt}
    		EfficientNetV2-L~\cite{tan2021efficientnetv2} & 121 & 52 & 85.7 \\
                ConvNext-L~\cite{convnext} & 198 & 34.4 & 84.3 \\
                ConvNext-L $\uparrow$384~\cite{convnext} & 198 & 101.0 & 85.5 \\
    		\Xhline{0.3pt}
    		Swin-B~\cite{swin} & 88 & 15.4 & 83.3 \\
    		Focal-B~\cite{focal} & 90 & 16.0 & 83.8 \\
    		CSwin-B~\cite{cswin} & 78 & 15.0 & 84.2 \\
    		\Xhline{0.3pt}
    		CoAtNet-3~\cite{dai2021coatnet} & 168 & 34.7 & 84.5 \\
    		LV-ViT-L 
 $\uparrow$288~\cite{lvvit} & 150 & 59.0 & 85.3 \\
    		UniFormer-L*~\cite{uniformer} & 100 & 12.6 & 85.6 \\
    		UniFormer-L* $\uparrow$384~\cite{uniformer} & 100 & 39.2 & 86.0 \\
    		MViTv2-L~\cite{mvitv2} & 218 & 42.1 & 85.3\\
                MViTv2-L $\uparrow$384~\cite{mvitv2} & 218 & 140.2 & 86.0\\
            \cellcolor{gray!20}{StructViT-L-4-1*} &  \cellcolor{gray!20}{103} &   \cellcolor{gray!20}{15.4} & \cellcolor{gray!20}{86.0} \\
            \cellcolor{gray!20}{StructViT-L-4-1* $\uparrow$384} &  \cellcolor{gray!20}{103} &   \cellcolor{gray!20}{85.2} & \cellcolor{gray!20}{\textbf{86.7}} \\
    	    \Xhline{1.0pt}
    	\end{tabular}
	}
        \vspace{-3mm}
	\caption{\textbf{Comparisons to the state-of-the-art methods on ImageNet-1K.} *Trained with Token Labeling \cite{lvvit}. 
    }
    \label{tab:sota_in1k}
    \vspace{-4mm}
\end{table}

\subsubsection{Image Classification}
In Table~\ref{tab:sota_in1k}, we compare StructViT with other state-of-the-art CNNs, ViTs, and their hybrid models.
The results show that StructViT outperforms other methods in all sizes.
Compared to EfficientNets~\cite{tan2019efficientnet,tan2021efficientnetv2} that are obtained by extensive architecture search, our models show comparable or even better performances in both base and large configurations, requiring much less amount of computational cost.
Compared to our baseline, UniFormers, StructViTs consistently improve top-1 accuracy regardless of their sizes, demonstrating the benefits of learning geometric structures in image understanding.
We further evaluate our models on stronger setup using Token Labeling~\cite{lvvit} and $384\times384$ images, and observe consistent improvements over the baselines.
While StructSA introduces some additional FLOPs, we also test variants whose stride for key/value convolutions is set to 2 (S-4-2 and B-4-2) matching its FLOPs to that of the baseline; 
we still observe some gain with the base model (B-4-2) without additional FLOPs while the small model (S-4-2) shows comparable results.
For more comprehensive analysis, we provide experimental results on dense prediction tasks, \ie, object detection, semantic segmentation, and instance segmentation.
Please refer to Appendix~\ref{sec:supp_dense_prediction} for the detail.

\subsubsection{Video Classification}
\noindent\textbf{Kinetics-400.}
Table~\ref{tab:sota_k400} compares our method with previous state-of-the-art methods on Kinetics-400.
Each block in the table groups methods based on their network structures: CNNs, ViTs, and hybrid methods.
We first observe that our best model (B-4-1) achieves state-of-the-art performance.
Our method outperforms CNN-based approaches even with less computational cost (S-4-2) in most cases.
Compared to MoViNets~\cite{kondratyuk2021movinets} which are the most advanced CNNs obtained by an extensive NAS, our method shows comparable scores with fewer FLOPs (S-4-1).

When we compare our model to the ViT-based ones, our model outperforms them by large margins while using significantly fewer compute.
For instance, StructViT-B-4-1 with single crop (second last row in Table~\ref{tab:sota_k400}) shows top-1 accuracy gain by 1.6\%p while using only 55\% of computes compared to MTV-B, the best performing ViT-based model.
Note also that our model is pretrained on ImageNet-1K, which is much smaller than ImageNet-21K on which the ViT-based models are pretrained.

Finally, our best models (S-8-1 \& B-4-1) show 0.5\%p to 0.8\%p accuracy gains over the baseline UniFormer models in different size configurations.
When we use larger strides (S-4-2 and B-4-2) to match the FLOPs of the baselines, we still observe accuracy gains ranging from 0.2\%p to 0.3\%p.

\begin{table}[t]
\centering
    \setlength\tabcolsep{1.6pt}
    \scalebox{0.9}{
    	\begin{tabular}{l|l|l|l|cc}
    	    \Xhline{1.0pt}
    		\multirow{2}*{method} & \multirow{2}*{pretrain} & frame$\times$ & \makecell[c]{FLOPs} & \multicolumn{2}{c}{K400}\\
    		 ~ & ~ & crop$\times$clip & \makecell[c]{(G)} & top-1 & top-5\\
    	    \Xhline{0.7pt}
    		SlowFast+NL~\cite{feichtenhofer2019slowfast} & - & 16$\times$3$\times$10 & 7020 & 79.8 & 93.9 \\
    		ip-CSN~\cite{tran2019video} & Sports1M & 32$\times$3$\times$10 & 3270 & 79.2 & 93.8 \\
    		X3D-XL~\cite{feichtenhofer2020x3d} & - & 16$\times$3$\times$10 & 1452 & 79.1 & 93.9 \\
    		MoViNet-A5~\cite{kondratyuk2021movinets} & - & 120$\times$1$\times$1 & 281 & 80.9 & 94.9  \\
    		MoViNet-A6~\cite{kondratyuk2021movinets} & - & 120$\times$1$\times$1 & 386 & 81.5 & 95.3  \\
    	    \Xhline{0.7pt}
    		TimeSformer-HR~\cite{timesformer} & IN-21K & 16$\times$3$\times$1 & 5109 & 79.7 & 94.4 \\
    		TimeSformer-L~\cite{timesformer} & IN-21K & 96$\times$3$\times$1 & 7140 & 80.7 & 94.7 \\
			X-ViT~\cite{xvit} & IN-21K & 16$\times$3$\times$1 & 850 & 80.2 & 94.7 \\
			Mformer-HR~\cite{mformer} & IN-21K & 16$\times$3$\times$10 & 28764 & 81.1 & 95.2 \\
    		ViViT-L~\cite{arnab2021vivit} & IN-21K & 16$\times$3$\times$4 & 17352 & 80.6 & 94.7 \\
    		Swin-B~\cite{videoswin} & IN-1K & 32$\times$3$\times$4 & 3384 & 80.6 & 94.6 \\
    		Swin-B~\cite{videoswin} & IN-21K & 32$\times$3$\times$4 & 3384 & 82.7 & 95.5 \\
            MTV-B~\cite{mtv} & IN-21K & 32$\times$3$\times$4 & 4790 & 81.8 & 95.0 \\
            \Xhline{0.7pt}
            MViT-B,16$\times$4~\cite{mvit} & - & 16$\times$1$\times$5 & 353 & 78.4 & 93.5 \\
    		MViT-B,32$\times$3~\cite{mvit} & - & 32$\times$1$\times$5 & 850 & 80.2 & 94.4 \\
    		Dualformer-S~\cite{liang2021dualformer} & IN-1K & 32$\times$1$\times$4 & 636 & 80.6 & 94.9 \\
            Dualformer-B~\cite{liang2021dualformer} & IN-1K & 32$\times$1$\times$4 & 1072 & 81.1 & 95.0 \\
            UniFormer-S~\cite{uniformer} & IN-1K & 16$\times$1$\times$4 & 167 & 80.8 & 94.7 \\
            UniFormer-B~\cite{uniformer} & IN-1K & 32$\times$1$\times$4 & 1036 & 82.9 & 95.4 \\
    		MViTv2-B,32$\times$3~\cite{mvitv2} & - & 32$\times$1$\times$5 & 1125 & 82.9 & 95.7 \\
    		\cellcolor{gray!20}{StructViT-S-4-2} & \cellcolor{gray!20}{IN-1K} & \cellcolor{gray!20}{16$\times$1$\times$4} & \cellcolor{gray!20}{169} & \cellcolor{gray!20}{81.1} & \cellcolor{gray!20}{95.5} \\
    		\cellcolor{gray!20}{StructViT-S-4-1} & \cellcolor{gray!20}{IN-1K} & \cellcolor{gray!20}{16$\times$1$\times$4} & \cellcolor{gray!20}{327} & \cellcolor{gray!20}{81.4} & \cellcolor{gray!20}{95.7} \\
    		\cellcolor{gray!20}{StructViT-S-8-1} & \cellcolor{gray!20}{IN-1K} & \cellcolor{gray!20}{16$\times$1$\times$4} & \cellcolor{gray!20}{541} & \cellcolor{gray!20}{81.6} & \cellcolor{gray!20}{95.8} \\
    		\cellcolor{gray!20}{StructViT-B-4-2} & \cellcolor{gray!20}{IN-1K} & \cellcolor{gray!20}{32$\times$1$\times$4} & \cellcolor{gray!20}{1045} & \cellcolor{gray!20}{83.1} & \cellcolor{gray!20}{95.5} \\
    		\cellcolor{gray!20}{StructViT-B-4-1} & \cellcolor{gray!20}{IN-1K} & \cellcolor{gray!20}{32$\times$1$\times$4} & \cellcolor{gray!20}{2658} & \cellcolor{gray!20}{83.3} & \cellcolor{gray!20}{95.6} \\
    		\cellcolor{gray!20}{StructViT-B-4-1} & \cellcolor{gray!20}{IN-1K} & \cellcolor{gray!20}{32$\times$3$\times$4} & \cellcolor{gray!20}{7974} & \cellcolor{gray!20}{\textbf{83.4}} & \cellcolor{gray!20}{\textbf{95.8}} \\
    	    \Xhline{1.0pt}
    	\end{tabular}
    }
    \vspace{-3mm}
    \caption{\textbf{Comparisons to the state-of-the-art methods on Kinetics-400.} 
    }
    \label{tab:sota_k400}
    \vspace{-4mm}
\end{table}

\begin{table*}[t]
\vspace{-2mm}
	\setlength\tabcolsep{4pt}
\begin{minipage}{0.66\linewidth}
    \begin{subtable}[t]{\columnwidth}
    \scalebox{0.9}{
    	\begin{tabular}{l|l|l|l|cc|cc}
    	    \Xhline{1.0pt}
    		\multirow{2}*{method} & \multirow{2}*{pretrain} & frame$\times$ & \makecell[c]{FLOPs} & \multicolumn{2}{c}{Something V1} & \multicolumn{2}{c}{Something V2} \\
    		 ~ & ~ & crop$\times$clip & \makecell[c]{(G)} & top-1 & top-5 & top-1 & top-5 \\
    	    \Xhline{0.7pt}
    		TEA~\cite{li2020tea} & IN-1K & 16$\times$1$\times$1 & 70  & 51.9 & 80.3 & - & - \\
    		MSNet~\cite{motionsqueeze} & IN-1K & 16$\times$1$\times$1 & 101  & 52.1 & 82.3 & 64.7 & 89.4 \\
    		CT-Net~\cite{li2021ctnet} & IN-1K & 16$\times$1$\times$1 & 75  & 52.5 & 80.9 & 64.5 & 89.3 \\
    		TDN~\cite{wang2021tdn} & IN-1K & 16$\times$1$\times$1 & 72  & 53.9 & 82.1 & 65.3 & 89.5 \\
    		SELFYNet~\cite{selfy} & IN-1K & 16$\times$1$\times$1 & 77 & 54.3 & 82.9 & 65.7 & 89.8 \\
    		RSANet~\cite{rsa} & IN-1K & 16$\times$1$\times$1 & 72 & 54.0 & 81.1 & 66.0 & 89.9 \\
    	    \Xhline{0.7pt}
    		TimeSformer-HR~\cite{timesformer} & IN-21K & 16$\times$3$\times$1 & 5109 & - & - & 62.5 & - \\
    		TimeSformer-L~\cite{timesformer} & IN-21K & 96$\times$3$\times$1 & 7140 & - & - & 62.3 & - \\
    		ViViT-L~\cite{arnab2021vivit} & K400 & 16$\times$3$\times$4 & 11892 & - & - & 65.4 & 89.8 \\
    		X-ViT~\cite{xvit} & IN-21K & 32$\times$3$\times$1 & 1270 & - & - & 65.4 & 90.7 \\
			Mformer-HR~\cite{mformer} & K400 & 16$\times$3$\times$1 & 2876 & - & - & 67.1 & 90.6 \\
			Mformer-L~\cite{mformer} & K400 & 32$\times$3$\times$1 & 3555 & - & - & 68.1 & 91.2 \\
    		Swin-B~\cite{videoswin} & K400 & 32$\times$3$\times$1 & 963 & - & - & 69.6 & 92.7 \\
    		\Xhline{0.7pt}
    		MViT-B,64$\times$3~\cite{mvit} & K400 & 64$\times$1$\times$3 & 1365 & - & - & 67.7 & 90.9 \\
    		MViT-B-24,32$\times$3~\cite{mvit} & K600 & 32$\times$1$\times$3 & 708 & - & - & 68.7 & 91.5 \\
    		UniFormer-S~\cite{uniformer} & K400 & 16$\times$3$\times$1 & 125 & 57.2 & 84.9 & 67.7 & 91.4 \\
    		UniFormer-B~\cite{uniformer} & K400 & 32$\times$3$\times$1 & 777 & 60.9 & 87.3 & 71.2 & 92.8 \\
    		MViTv2-B~\cite{mvitv2} & K400 & 32$\times$3$\times$1 & 675 & - & - & 70.5 & 92.7 \\
    	    \cellcolor{gray!20}{StructViT-S-4-2} & \cellcolor{gray!20}{K400} & \cellcolor{gray!20}{16$\times$3$\times$1} & \cellcolor{gray!20}{126} & \cellcolor{gray!20}{57.6} & \cellcolor{gray!20}{85.3} & \cellcolor{gray!20}{68.3} & \cellcolor{gray!20}{91.3} \\
    		\cellcolor{gray!20}{StructViT-S-4-1} & \cellcolor{gray!20}{K400} & \cellcolor{gray!20}{16$\times$3$\times$1} & \cellcolor{gray!20}{246} & \cellcolor{gray!20}{58.0} & \cellcolor{gray!20}{85.5} & \cellcolor{gray!20}{68.8} & \cellcolor{gray!20}{91.9} \\
    		\cellcolor{gray!20}{StructViT-S-8-1} & \cellcolor{gray!20}{K400} & \cellcolor{gray!20}{16$\times$3$\times$1} & \cellcolor{gray!20}{405} & \cellcolor{gray!20}{58.0} & \cellcolor{gray!20}{85.7} & \cellcolor{gray!20}{69.0} & \cellcolor{gray!20}{92.1} \\
            \cellcolor{gray!20}{StructViT-B-4-2} & \cellcolor{gray!20}{K400} & \cellcolor{gray!20}{32$\times$3$\times$1} & \cellcolor{gray!20}{784} & \cellcolor{gray!20}{61.1} & \cellcolor{gray!20}{87.7} & \cellcolor{gray!20}{71.1} & \cellcolor{gray!20}{92.7} \\
            \cellcolor{gray!20}{StructViT-B-4-1} & \cellcolor{gray!20}{K400} & \cellcolor{gray!20}{32$\times$3$\times$1} & \cellcolor{gray!20}{1963} & \cellcolor{gray!20}{\textbf{61.3}} & \cellcolor{gray!20}{\textbf{87.8}} & \cellcolor{gray!20}{\textbf{71.5}} & \cellcolor{gray!20}{\textbf{93.1}} \\
    	    \Xhline{1.0pt}
    	\end{tabular}
    }
    \caption{\textbf{Something-Something V1 \& V2}}
    \label{tab:sota_something}
    \end{subtable}
\end{minipage}
\begin{minipage}{0.33\linewidth}
\begin{subtable}[t]{\columnwidth}
    \centering  
    \captionsetup{width=\columnwidth}
    \scalebox{0.9}{
    \begin{tabular}[t]{L{4.1cm}|C{1.3cm}}
    \Xhline{1.0pt}
    model & top-1 \\
    \Xhline{0.7pt}
    SlowFast-R101~\cite{feichtenhofer2019slowfast} &  77.6 \\
    TimeSformer~\cite{timesformer}      & 75.0 \\
    TimeSformer-HR~\cite{timesformer}   & 78.0 \\
    SViT-DD~\cite{ben2022bringing} & 79.8 \\
    TimeSformer-L~\cite{timesformer}    & 81.0 \\
    TQN~\cite{zhang2021temporal} & 81.8 \\
    RSANet-R50~\cite{rsa} & 84.2 \\
    UniFormer-B*~\cite{uniformer}    & 87.4 \\
    \textcolor{gray}{ORViT$^\dagger$}~\cite{herzig2022object} & \textcolor{gray}{88.0} \\
    \cellcolor{gray!20}{StructViT-B-4-2} & \cellcolor{gray!20}{87.8}  \\
    \cellcolor{gray!20}{StructViT-B-4-1} & \cellcolor{gray!20}{\textbf{88.3}}  \\
    \Xhline{1.0pt}
    \end{tabular}
    }
    \caption{\textbf{Diving-48}} 
    \label{tab:sota_diving}
\end{subtable}

\vspace{10mm}

\begin{subtable}[t]{\columnwidth}
    \captionsetup{width=\columnwidth}
    \centering
    \scalebox{0.9}{
    \begin{tabular}[t]{L{2.75cm}|C{1.25cm}C{1.15cm}}             
    \Xhline{1.0pt}
    model & Gym288 & Gym99 \\
    \Xhline{0.7pt}
    TRN~\cite{zhou2018temporal}            & 33.1 & 68.7 \\
    I3D~\cite{carreira2017quo}             & 27.9 & 63.2 \\
    TSM~\cite{lin2019tsm}                  & 34.8 & 70.6 \\
    TSM$_{\textrm{Two-stream}}$~\cite{lin2019tsm}  & 46.5 & 81.2 \\
    RSANet-R50~\cite{rsa}                       & 50.9 & 86.4 \\
    UniFormer-B*~\cite{uniformer}    & 53.5 & 88.9 \\
    \cellcolor{gray!20}{StructViT-B-4-2} & \cellcolor{gray!20}{53.8} & \cellcolor{gray!20}{89.3} \\
    \cellcolor{gray!20}{StructViT-B-4-1} & \cellcolor{gray!20}{ \textbf{54.2}} & \cellcolor{gray!20}{\textbf{89.5}} \\
    \Xhline{0.7pt}
    \end{tabular}
    }
    \caption{\textbf{FineGym}
    }
    \label{tab:sota_finegym}
\end{subtable}
\end{minipage}
      \vspace{-2mm}
    \caption{\textbf{Comparisons to the state-of-the-art methods on three motion-centric video classification benchmarks.} 
    Our StructViT achieves new state-of-the-art on all the benchmarks.
    For FineGym, we measure averaged per-class accuracy while top-$k$ accuracy is measured for Something-Something and Diving-48.
    *Reproduced by our experimental setup. $\dagger$Trained with additional bbox annotations.}
    \label{tab:sota_motion}
    \vspace{-2mm}
\end{table*}

\begin{figure*}[t]
  \scalebox{0.9}{
      \includegraphics[width=\linewidth]{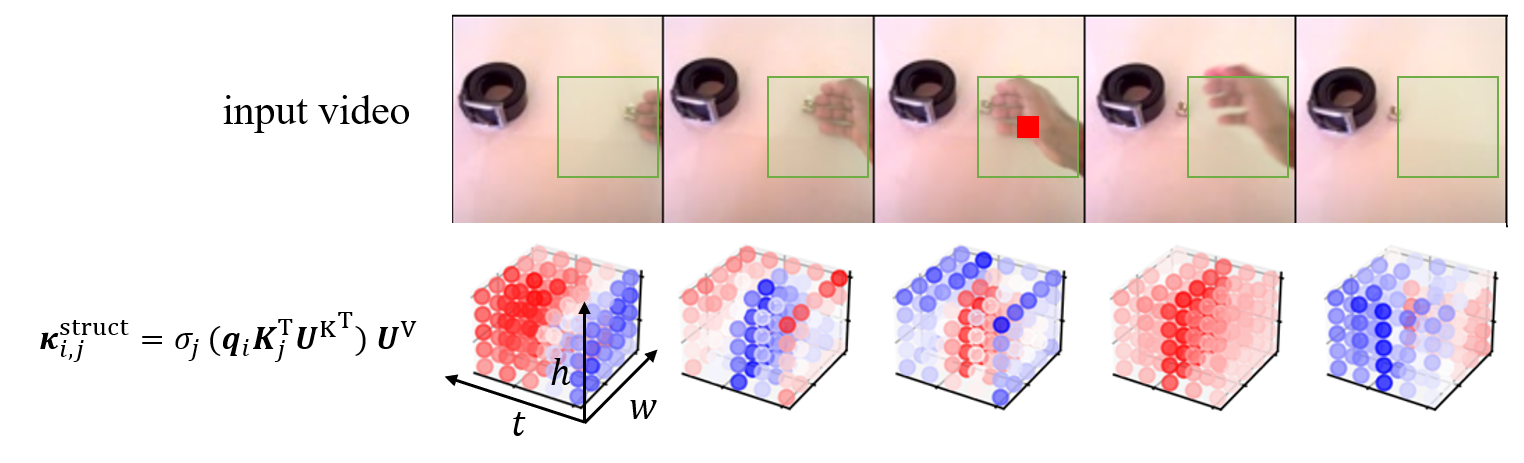}}
      \vspace{-4mm}
      \caption{\textbf{Visualization of dynamic kernels $\bm\kappa^{\mathrm{struct}}_{i,j}$ in StructSA on Something-Something-V1.} 
      The top row shows the input frames that contain the input spatiotemporal local context (indicated by green boxes) used in the dynamic kernel computation.
      The bottom row presents the resulting dynamic kernels $\bm\kappa^\mathrm{struct}_{i,j}$ for a StructSA head when $i=j$.
      Note that the computed dynamic kernels are computed with self-similarity map ($i=j$) to illustrate its effectiveness in capturing motions in videos.
      We use StructViT-S-4-1 with $M=5 \times 5 \times 5$.
    }
      \label{fig:kernel_visualization}
      \vspace{-2mm}
\end{figure*}

\noindent\textbf{Something-Something, Diving-48 and FineGym.}
Table~\ref{tab:sota_something} summarizes the results on Something-Something-V1\&V2.
We observe the same trends as on Kinetics-400.
StructViT-S-4-2 outperforms UniFormer-S on Something-Something-V2 by 0.6\%p in top-1 accuracy, and StructViT-S-8-1 enlarges the gap to 1.3\%p, leveraging correlation structures more effectively.
StructViT-B-4-1 achieves new state-of-the-art performances on both V1 and V2 without the strong data augmentation methods used in \cite{uniformer, mvitv2}.

Table~\ref{tab:sota_diving} and Table~\ref{tab:sota_finegym} show the results on Diving-48~\cite{li2018resound} and FineGym~\cite{shao2020finegym}.
Our models outperform the baseline, UniFormer-B, obtaining significant accuracy gains by 0.9\%p and 0.7\%p on Diving-48 and FineGym, respectively.
This indicates that learning spatio-temporal correlation structures play a crucial role in capturing fine-grained motion patterns.
Our model sets new state-of-the-art performances with large margins (4.1\%p on Diving-48; 3.3\%p and 3.1\%p on FineGym) over the previous methods without additional box annotations on both datasets.
Note that ORViT~\cite{herzig2022object} uses additional object bounding box annotations to train an object detector.

\subsection{Visualizations of StructSA}
Figure~\ref{fig:kernel_visualization} visualizes example dynamic kernels $\bm\kappa^{\mathrm{struct}}_{i,j}$ computed from self-similarity map ($i=j$) on Something-Something-V1 to illustrate how StructSA encodes motion features from the spatiotemporal correlation structure.
We observe that StructSA builds kernels for spatiotemporal gradient filters similar to those that are already known to be effective for capturing different types of motions~\cite{szeliski2010computer}, \eg, Sobel filters (first) or Laplacian filters (second and third), over local contexts similarly to \cite{rsa}.

\section{Conclusion}
\vspace{-2mm}
We have introduced a novel self-attention mechanism, StructSA, that exploits rich structural patterns of query-key correlation for visual representation learning.
StructSA leverages spatial (and temporal) structures of local correlations and aggregates chunks of local features globally across entire locations. %
 Structural Vision Transformer (StructViT) using StructSA as the main attention module achieves state-of-the-art results on both image and video classification benchmarks.
 We believe leveraging structural patterns of correlation in attention will also benefit other tasks in computer vision and natural language processing. We leave this for future work.

\vspace{-4mm}
\paragraph{Acknowledgements}
This work was supported by the IITP grants (No. 2022-0-00290: Visual Intelligence for space-time understanding and generation (45\%), No. 2022-0-00264: Comprehensive video understanding and
generation with knowledge-based deep logic (50\%), No. 2019-0-01906: AI graduate school program at POSTECH (5\%)) funded by Ministry of Science and ICT, Korea.

{\small
    \bibliographystyle{ieeenat_fullname}
    \bibliography{reference}
}
\clearpage
\appendix

\section{Full Derivation}
\label{sec:supp_full_derivation}
\subsection{Structural Self-Attention}
In Sec.~\ref{sec:structsa} of our main paper, we explain StructSA with a slightly simpler version that uses dot-product correlation. 
Here we provide the full version of StructSA used in our experiments, which captures fine-grained correlation structures by employing channel-wise correlation.
\paragraph{Channel-wise Correlation.}
Since the dot product correlation $\vq_i \mK^\T$ reduces all the channels of the query and the keys, it might lose rich semantic information.
We instead use the Hadamard product~\cite{pirsiavash2009bilinear,kim2016hadamard,rsa} to leverage richer channel-wise correlation structures for generating the final attention weights.
SQKA extracts structural patterns from the channel-wise correlation applying convolution as 
\begin{align}
    \mA_i &= \sigma \left( \texttt{conv} \left( \mathrm{diag}(\vq_i) \mK, \tH^{\mathrm{K}} \right) \right) \in \mathbb{R}^{N \times D}, \label{eq:chnlwise_sqka} \\
    \tH^{\mathrm{K}} &= [ \mH^{\mathrm{K}}_1, \cdots, \mH^{\mathrm{K}}_D ] \in \mathbb{R}^{D \times M \times C},
\end{align}
where $\mathrm{diag}(\cdot)$ is a function that outputs a square diagonal matrix from an input vector and $\tH^{\mathrm{K}}$ represents $D$ convolutional filters of which kernel size and input channel are $M$ and $C$, respectively.
Each score of $\mA_i$ is computed as
\begin{align}
    \va_{i,j} &= \sigma_j \left( \mathrm{vec}\left( \mathrm{diag}\left(\vq_i\right)  \mK_j \right) {f(\tH^{\mathrm{K}})}^\T \right), \label{eq:chnlwise_sqka_element}\\
    f(\tH^{\mathrm{K}}) &= [ \mathrm{vec}(\mH^{\mathrm{K}}_1), \cdots, \mathrm{vec}(\mH^{\mathrm{K}}_D) ] \in \mathbb{R}^{D \times MC},
\end{align}
where $\mathrm{vec}(\cdot)$ is a vectorization function.
Compared to $\mU^{\mathrm{K}}$ where each column takes a single correlation map to detect a structural pattern, each of $f(\tH^{\mathrm{K}})$, \ie, $\mathrm{vec}(\mH^{\mathrm{K}}_d)$, extracts a pattern from the whole $C$ correlation maps using fine-grained channel-wise correlation.
One potential drawback of the channel-wise correlation map, $\mathrm{diag}(\vq_i) \mK$, would be to increase the memory complexity $C$-times larger compared to that of dot-product correlation map, \ie, $\mathcal{O}(N^2C)$ vs. $\mathcal{O}(N^2)$.
To address the issue, we permute the computation orders of Eq.~(\ref{eq:chnlwise_sqka_element}) as
\begin{align}
    \va_{i,j} &= \sigma_j \left( \sum^{C}_{c=1} \sum^{M}_{m=1} (\vq_i)_c (\mK_j)_{m,c} (\tH^{\mathrm{K}})_{:,m,c}  \right) \nonumber\\
    &= \sigma_j \left( \sum^{C}_{c=1} (\vq_i)_c \sum^{M}_{m=1} (\mK_j)_{m,c} (\tH^{\mathrm{K}})_{:,m,c}  \right) \nonumber\\
    &= \sigma_j \left( \vq_i (\mK_j \ast \tH^{\mathrm{K}})  \right).
\end{align}
where we first compute $\mK_j \ast \tH^{\mathrm{K}}$, which requires memory complexity of $\mathcal{O}(NCD)$, and then multiply it with $\vq_i$. This enables us to compute $\va_{i,j}$ in a memory-efficient way when $N>D$ without explicit computation of channel-wise correlation maps.
\paragraph{Channel-wise Context Value Aggregation.}
We also extend the context aggregators $\mU^{\mathrm{V}}$, which are shared by different channels, to be channel-wise aggregators, so that they can learn aggregation weights more adaptive to each channel of the values.
Each channel of StructSA output is computed as
\begin{align}
    (\vy_i)_c &= \sum_{j=1}^N \sigma_j \left( \mathrm{vec}\left( \mathrm{diag}(\vq_i) \mK_j \right) {f(\tH^{\mathrm{K}})}^\T \right) (\tH^{\mathrm{V}})_{:,:,c} (\mV_j)_{:,c}, \\
    \tH^{\mathrm{V}} &= [ \mH^{\mathrm{V}}_1, \cdots, \mH^{\mathrm{V}}_D ] \in \mathbb{R}^{D \times M \times C}.
\end{align}
Compared to $\mU^{\mathrm{V}}$ that produces a single kernel shared by every channel, $(\tH^{\mathrm{V}})_{:,:,c}$ generates $C$ different spatial kernels aggregating the context with diverse patterns.
We conduct experiments to investigate the effect of channel-wise correlation and context aggregation in Appendix~\ref{sec:supp_ablation}.
We use this version of StructSA as a basic operation in our main paper.

\subsection{Convolutional Self-Attention}
For the sake of simplicity in derivation, ConvSA (Eqs.~(\ref{eq:conv_proj_key})-(\ref{eq:convolutional_self_attention})) in Sec.~\ref{sec:structure_convsa} is described as sharing channel-wise convolution weights across channels for key and value projection, which is not exactly the same as those used in previous ConvSA-based methods~\cite{cvt,mvit,mvitv2}.
We here provide a full derivation of ConvSA with conventional channel-wise convolution, of which weights are not shared across channels.
Given the channel-wise convolution weights $\mH^{\mathrm{K}}, \mH^{\mathrm{V}} \in \mathbb{R}^{M \times C}$, $c$-th channel of each key $k_{i}^{\mathrm{conv}}$ and value $v_{i}^{\mathrm{conv}}$ is computed as
\begin{align}
    \vspace{-2mm}
    (\vk^{\mathrm{conv}}_i)_{c} &= ({\mH^{\mathrm{K}}}^\T)_c (\mK_i)_{:,c}  \in \mathbb{R}, \label{eq:conv_proj_key_apdx} \\
    (\vv^{\mathrm{conv}}_i)_{c} &= ({\mH^{\mathrm{V}}}^\T)_c (\mV_i)_{:,c}  \in \mathbb{R}, \label{eq:conv_proj_value_apdx}
\end{align}
where $(\mK_i)_{:,c},(\mV_i)_{:,c} \in \mathbb{R}^{M \times 1}$ indicate features in the $c$-th channel of $(\mK_i)$ and $(\mV_i)$, respectively.
Plugging Eqs.~(\ref{eq:conv_proj_key_apdx}) and (\ref{eq:conv_proj_value_apdx}) into Eq.~(\ref{eq:convolutional_self_attention}), each channel of ConvSA output is computed as
\vspace{-2mm}
\begin{align}
    (\vy_i)_c &= \sum_{j=1}^N \sigma_j \left( \vq_i {\vk_j^{\mathrm{conv}}}^\mathsf{T}  \right) (\vv_j^{\mathrm{conv}})_c \nonumber\\
    &= \sum_{j=1}^N \sigma_j \left( \sum^{C}_{c=1} (\vq_i)_c \left( ({\mH^{\mathrm{K}}}^\T)_c (\mK_j)_{:,c} \right)  \right) (\vv_j^{\mathrm{conv}})_c \nonumber\\
    &= \sum_{j=1}^N \sigma_j \left( \sum^{C}_{c=1} \sum^{M}_{m=1} (\vq_i)_c (\mK_j)_{m,c} (\mH^{\mathrm{K}})_{m,c} \right) (\vv_j^{\mathrm{conv}})_c \nonumber\\
    &= \sum_{j=1}^N \sigma_j \left( \mathrm{vec} \left( \mathrm{diag}(\vq_i) \mK_j \right) \mathrm{vec} \left(\mH^{\mathrm{K}} \right) \right) ({\mH^{\mathrm{V}}}^\T)_c (\mV_j)_{:,c}.
    \label{eq:convsa_apdx}
\end{align}
This reveals that the channel-wise convolution weights $\mH^{\mathrm{K}}$ for the key projection, in fact, act as a \textit{pattern detector} that extracts a single structural pattern from the channel-wise correlation, while those of $\mH^{\mathrm{V}}$ perform as a \textit{channel-wise context aggregator} that generates a spatial kernel weights for every channel.
Despite the capability of capturing a channel-wise correlation structure, it still learns a single pattern only from the rich channel-wise correlation, thus being limited in leveraging diverse structural patterns for the attention weight generation compared to StructSA.

\section{Additional Ablation Experiments}
\label{sec:supp_ablation}
Here we provide additional ablation experiments to validate design components in StructSA.
We follow the same training and testing protocols in Sec.~\ref{sec:experimental_setup} of our main paper.

\paragraph{Channel-wise Correlation and Aggregation.}
Table~\ref{tab:supp_ablation_channelwise} summarizes the effectiveness of the channel-wise correlation and aggregation.
Compared to the dot-product correlation, the channel-wise correlation improves the top-1 accuracy by 0.3\%p and 1.2\%p on ImageNet-1K and Something-Something V1 datasets, respectively, validating that fine-grained structures from the channel-wise correlation are beneficial to the attention weight generation.
As we use channel-wise context aggregator, we obtain additional improvements by 0.1\%p and 0.5\%p on both datasets.
\vspace{-3mm}
\paragraph{Different Combinations of $\mU^{\mathrm{K}}$ and $\mU^{\mathrm{V}}$.}
In Table~\ref{tab:supp_ablation_m}, we investigate different combinations of $\mU^{\mathrm{K}}$ and $\mU^{\mathrm{V}}$ varying the size of the kernel size $M$.
As discussed in Sec.4.3, using the large kernel size $M$ on both $\mU^{\mathrm{K}}$ and $\mU^{\mathrm{V}}$ improves the performance, demonstrating the effectiveness of SQKA and contextual aggregation.
The performance saturates as $M$ gets larger than $5 \times 5 \times 5$.
We set the kernel size $M$ of $\mU^{\mathrm{K}}$ and $\mU^{\mathrm{V}}$ to $3\times3\times3$ as default considering computation-accuracy trade-off.

\begin{table}[t]
\setlength\tabcolsep{3pt}

\begin{subtable}[t]{\columnwidth}
    \centering  
    \captionsetup{width=\columnwidth}
    \scalebox{0.9}{
    \begin{tabular}[t]{C{2.0cm}C{2.0cm}|C{0.8cm}C{0.8cm}|C{0.8cm}C{0.8cm}}
    \Xhline{1.0pt}
    \multicolumn{2}{c|}{channel-wise} & \multicolumn{2}{c|}{ImageNet-1K} & \multicolumn{2}{c}{Something V1} \\
    correlation & aggregation & top-1 & top-5 & top-1 & top-5 \\
    \Xhline{0.7pt}
    &  & 80.7 & 95.1 & 48.7 & 77.2 \\
    \checkmark &  & 81.0 & 95.3 & 49.9 & 77.7 \\
    \checkmark & \checkmark & \textbf{81.1} & \textbf{95.4} & \textbf{50.4} & \textbf{78.2} \\
    \Xhline{1.0pt}
    \end{tabular}
    }
    \caption{\textbf{Channel-wise correlation and aggregation}. 
    }
    \label{tab:supp_ablation_channelwise}
\end{subtable}

\vspace{2mm}

\begin{subtable}[t]{\columnwidth}
    \centering  
    \captionsetup{width=\columnwidth}
    \scalebox{0.9}{
    \begin{tabular}[t]{C{2.0cm}C{2.0cm}|C{0.8cm}C{0.8cm}|C{0.8cm}C{0.8cm}}
    \Xhline{1.0pt}
    \multicolumn{2}{c|}{$M$} & \multicolumn{2}{c|}{ImageNet-1K} & \multicolumn{2}{c}{Something V1} \\
    $\mU^{\mathrm{K}}$& $\mU^{\mathrm{V}}$ & top-1 & top-5 & top-1 & top-5 \\
    \Xhline{0.7pt}
    - & - & 80.5 & 95.0 & 48.3 & 76.6 \\
    \Xhline{0.3pt}
    $3 \times 3 ~(\times 3)$ & $3 \times 3 ~(\times 3)$ & \underline{81.1} & \underline{95.4} & 50.4 & \textbf{78.2} \\
    \Xhline{0.3pt}
    $1 \times 1 ~(\times 1)$ & $3 \times 3 ~(\times 3)$ & 80.8 & 95.1 & 48.7 & 77.1 \\
    $5 \times 5 ~(\times 5)$ & $3 \times 3 ~(\times 3)$ & 81.0 & 95.3 & \textbf{50.6} & 78.1 \\
    $7 \times 7 ~(\times 7)$ & $3 \times 3 ~(\times 3)$ & 80.9 & 95.1 & 50.3 & 78.1 \\
    \Xhline{0.3pt}
    $3 \times 3 ~(\times 3)$ & $1 \times 1 ~(\times 1)$ & 80.9 & 95.2 & 49.6 & 77.5 \\
    $3 \times 3 ~(\times 3)$ & $5 \times 5 ~(\times 5)$ & \textbf{81.2} & \textbf{95.6} & 50.3 & \textbf{78.2} \\
    $3 \times 3 ~(\times 3)$ & $7 \times 7 ~(\times 7)$ & 81.0 & 95.4 & 50.3 & 77.8\\
    \Xhline{0.3pt}
    $1 \times 1 ~(\times 1)$ & $1 \times 1 ~(\times 1)$ & 80.6 & 95.0 & 48.5 & 76.9 \\
    $5 \times 5 ~(\times 5)$ & $5 \times 5 ~(\times 5)$ & \underline{81.1} & \underline{95.4} & \underline{50.5} & \textbf{78.2} \\
    $7 \times 7 ~(\times 7)$ & $7 \times 7 ~(\times 7)$ & 81.0 & 95.2 & \underline{50.5} & 78.1 \\
    \Xhline{1.0pt}
    \end{tabular}
    }
    \caption{\textbf{Kernel size $M$.} 
    }
    \label{tab:supp_ablation_m}
\end{subtable}
\vspace{-2mm}
\caption{\textbf{Ablation studies on ImageNet-1K and Something-Something V1}. Top-1 and top-5 accuracies (\%) are shown.
In Table~\ref{tab:supp_ablation_channelwise}, we set $D=4$, $M=3\times3\times3$. In Table~\ref{tab:supp_ablation_m}, we set $D=4$ as default. \textbf{Bold-faced} and \underline{underlined} numbers indicate the first and second highest scores, respectively.
}
\end{table}

\begin{table}[t]
\setlength\tabcolsep{1.0pt}
\centering  
\captionsetup{width=\columnwidth}
\scalebox{0.80}{
\begin{tabular}[t]{l|cccc|cccc}
\Xhline{1.0pt}
\multirow{2}{*}{method (DeiT-S)} & \multicolumn{4}{c|}{IN-1K} & \multicolumn{4}{c}{SS-V1} \\
& param & FLOPs & top-1 & top-5 & param & FLOPs & top-1 & top-5 \\
\Xhline{0.7pt}
ConvSA & 22.1M & 4.6G & 80.8 & 95.2 & 22.8M & 57.3G & 49.7 & 77.6 \\
ConvSA + channel$\uparrow$ & 27.9M & 5.8G & 80.9 & 95.2 & 34.3M & 80.4G & 49.9 & 77.9 \\
ConvSA + layer$\uparrow$ & 29.3M & 6.1G & 81.0 & \textbf{95.4} & 31.8M & 80.8G & 50.1 & 77.8 \\
\cellcolor{gray!20}{StructSA}  & \cellcolor{gray!20}{22.4M} & \cellcolor{gray!20}{5.7G}& \cellcolor{gray!20}{\textbf{81.1}} & \cellcolor{gray!20}{\textbf{95.4}} & \cellcolor{gray!20}{23.1M} & \cellcolor{gray!20}{80.4G} & \cellcolor{gray!20}{\textbf{50.4}} & \cellcolor{gray!20}{\textbf{78.2}} \\
\Xhline{1.0pt}
\end{tabular}
}
\vspace{-3mm}
\caption{\textbf{Comparison to ConvSA variants with similar FLOPs.}}
\label{tab:supp_comparison_convsa}
\vspace{-4mm}
\end{table}

\vspace{-3mm}
\paragraph{Comparison to ConvSA.}
Table~\ref{tab:supp_comparison_convsa} compares our StructSA to its ConvSA counterpart with a matching capacity, \ie, a similar number of parameters; we match their capacities by varying the number of channels or layers of the ConvSA backbone (DeiT-S). Our method achieves better accuracy-compute trade-off on ImageNet-1K and Something-Something V1 datasets. For example, StructSA outperforms the ConvSA variants with more parameters and compute on Something-Something V1 where learning motion dynamics may be more important for classification.%

\section{Results on Dense Prediction Tasks}
\label{sec:supp_dense_prediction}
We evaluate the generalizability of StructViT on various dense prediction tasks: object detection and instance segmentation on COCO 2017~\cite{coco} as well as semantic segmentation on ADE20K~\cite{ade20k}.
For object detection and instance segmentation, we use the Mask R-CNN~\cite{maskrcnn} with Hourglass UniFormer-\{S, B\}$_{h14}$~\cite{uniformer} as the backbone and then replace all SA blocks with our StructSA blocks.
We train the models for 12 epochs following the $1\times$ schedule in \cite{uniformer}.
Similarly, for semantic segmentation, we integrate StructSA blocks into Semantic FPN~\cite{semanticfpn} with Hourglass UniFormer-\{S,B\}$_{h32}$ backbone and train the models for 80K iterations following the protocols in \cite{pvt}.

Tables~\ref{tab:supp_coco} and \ref{tab:supp_ade20k} show consistent performance improvements across all benchmarks, affirming the effectiveness of StructSA. 
Specifically, in Table~\ref{tab:supp_coco}, StructViT-S-4-1$_{h14}$ outperforms the baseline UniFormer-S$_{h14}$ on both detection and segmentation tasks by 1.0 box mAP and 0.8 mask mAP, respectively.
Furthermore, semantic segmentation results in Table~\ref{tab:supp_ade20k} also shows the significant increase of mIOU over the baseline by 0.7\%p and 0.8\%p at both small and base scales, respectively.
These consistent improvements effectively demonstrate the generalizability of StructSA across various backbone scales and downstream tasks.

\section{Attention Map Visualization}
\label{sec:supp_visualization}
\begin{figure*}[t]
  \centering
    \begin{subfigure}{1\textwidth} %
    \refstepcounter{subfigure}\label{fig:supp_image_visualization:a}
    \refstepcounter{subfigure}\label{fig:supp_image_visualization:b}
    \refstepcounter{subfigure}\label{fig:supp_image_visualization:c}
    \refstepcounter{subfigure}\label{fig:supp_image_visualization:d}
    \refstepcounter{subfigure}\label{fig:supp_image_visualization:e}
    \refstepcounter{subfigure}\label{fig:supp_image_visualization:f}
    \end{subfigure}
   \includegraphics[width=\linewidth]{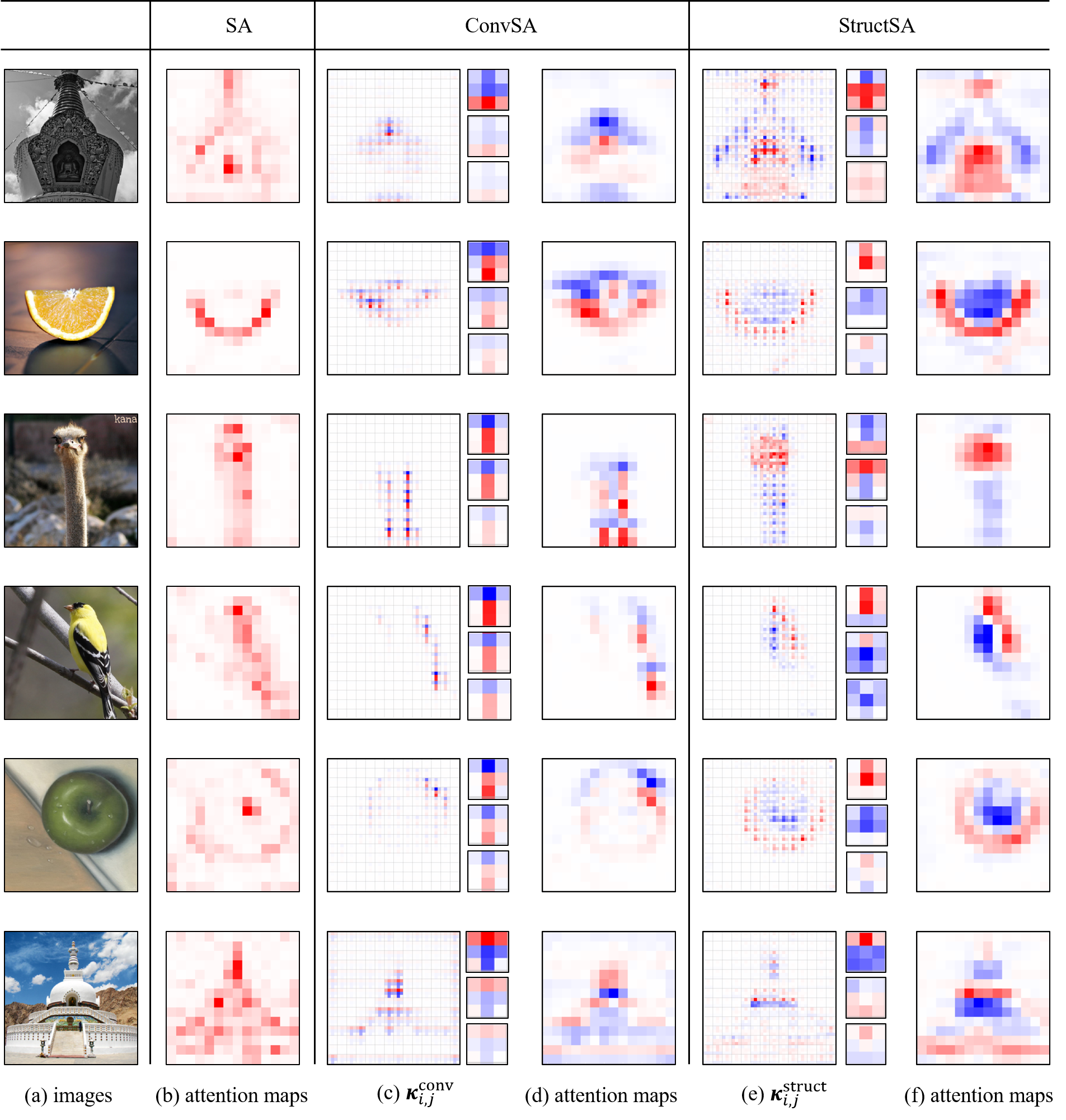}
      \caption{\textbf{Attention map visualization of SA, ConvSA, and StructSA on ImageNet-1K.}
      The query location $i$ is set to the center of the image and the kernel size $M=3 \times 3$.
      Given (a) input images, we illustrate (b) attention maps of SA, (c) dynamic kernels $\bm{\kappa}^{\mathrm{conv}}_{i,j}$, (d) final attention maps of ConvSA, \ie, aggregated weights of $\bm{\kappa}^{\mathrm{conv}}_{i,j}$, (e) dynamic kernels $\bm{\kappa}^{\mathrm{struct}}_{i,j}$, and (f) final attention maps of StructSA, \ie, aggregated weights of $\bm{\kappa}^{\mathrm{struct}}_{i,j}$, respectively.
      Note that in (c) and (e), each location $j$ has an aggregation map of the kernel size $M=3 \times 3$ and thus we show enlarged images for three different sampled locations $j$.
    }
      \label{fig:supp_image_visualization}
\end{figure*}

We visualize attention maps of SA, ConvSA, and StructSA to provide an in-depth comparison across the methods.
Different from SA, which uses individual query-key correlation as an attention weight for a single value feature (Fig.~\ref{fig:supp_image_visualization:b}), ConvSA and StructSA aggregate a local chunk of value features by generating dynamic kernels for each location.
ConvSA generates the dynamic kernels $\bm{\kappa}^{\mathrm{conv}}_{i,j}$, where spatial patterns are identical for all locations except for their scales (Fig.~\ref{fig:supp_image_visualization:c}).
In contrast, StructSA constructs the dynamic kernels $\bm{\kappa}^{\mathrm{struct}}_{i,j}$ in diverse aggregation patterns (Fig.~\ref{fig:supp_image_visualization:e}) by combining $D$ correlation pattern scores and context aggregation patterns as explained in Sec.~\ref{sec:structsa}.
This property of StructSA enables the model to effectively leverage geometric structures for visual representation learning.
To better observe the effect, we visualize the final attention maps of StructSA in Fig.~\ref{fig:supp_image_visualization:f} by spatially merging the overlapped kernels $\bm{\kappa}^{\mathrm{struct}}_{i,j}$ following the equation:
\begin{align}
    c^{\mathrm{struct}}_{i,j} = \sum^{M}_{m=0} ({\bm{\kappa}^{\mathrm{struct}}_{i,j-\lfloor M/2 \rfloor+m}})_m.
    \label{eq:supp_attention_map}
\end{align}

\begin{table}[t]
\setlength\tabcolsep{1.8pt}
\centering
\scalebox{0.90}{
\begin{tabular}[t]{l|c|ccc|ccc}
\Xhline{1.0pt}
\multirow{2}*{method} & \#param & \multicolumn{6}{c}{Mask R-CNN 1$\times$} \\
& (M) & AP$^\textrm{b}$ & AP$^\textrm{b}_{50}$ & AP$^\textrm{b}_{75}$ & AP$^\textrm{m}$ & AP$^\textrm{m}_{50}$ & AP$^\textrm{m}_{75}$ \\
\Xhline{0.7pt}
R50~\cite{he2015delving} & 44 & 38.0 & 58.6 & 41.4 & 34.4 & 55.1 & 36.7 \\
PVT-M~\cite{pvt} & 44 & 40.4 & 62.9 & 43.8 & 37.8 & 60.1 & 40.3 \\
Focal-T~\cite{focal} & 49 & 44.8 & 67.7 & 49.2 & 41.0 & 64.7 & 44.2 \\
PVTv2-B2~\cite{pvtv2} & 45 & 45.3 & 67.1 & 49.6 & 41.2 & 64.2 & 44.4 \\
UniFormer-S$_{h14}$~\cite{uniformer}  & 41 & 45.6 & 68.1 & 49.7 & 41.6 & 64.8 & 45.0 \\
\cellcolor{gray!20}{StructViT-S-4-1$_{h14}$} & \cellcolor{gray!20}{42} & \cellcolor{gray!20}{\textbf{46.6}}  & \cellcolor{gray!20}{\textbf{69.2}}  & \cellcolor{gray!20}{\textbf{51.5}} & \cellcolor{gray!20}{\textbf{42.8}} & \cellcolor{gray!20}{\textbf{65.5}} & \cellcolor{gray!20}{\textbf{46.1}}\\
\Xhline{0.7pt}
R101~\cite{he2015delving} & 63 & 40.4 & 61.1 & 44.2 & 36.4 & 57.7 & 38.8 \\
X101-32~\cite{resnext} & 63 & 41.9 & 62.5 & 45.9 & 37.5 & 59.4 & 40.2 \\
PVT-M~\cite{pvt} & 64 &  42.0 & 64.4 & 45.6 & 39.0 & 61.6 & 42.1 \\
PVT-L~\cite{pvt} & 81 & 42.9 & 65.0 & 46.6 & 39.5 & 61.9 & 42.5 \\
Twins-B~\cite{twins} & 76 & 45.2 & 67.6 & 49.3 & 41.5 & 64.5 & 44.8 \\
Swin-S~\cite{swin} & 69 & 44.8 & 66.6 & 48.9 & 40.9 & 63.8 & 44.2 \\
Swin-B~\cite{swin} & 107 & 46.9 & - & - & 42.3 & - & - \\
Focal-S~\cite{focal} & 71 & 47.4 & 69.8 & 51.9 & 42.8 & 66.6 & 46.1 \\
Focal-B~\cite{focal} & 110 & 47.8 & - & - & 43.2 & - & - \\
PVTv2-B5~\cite{pvtv2} & 101 & 47.4 & 68.6 & 51.9 & 42.5 & 65.7 & 6.0 \\
UniFormer-B$_{h14}$~\cite{uniformer} & 69 & 47.4 & 69.7 & 52.1 & 43.1 & 66.0 & 46.5 \\
\cellcolor{gray!20}{StructViT-B-4-1$_{h14}$} & \cellcolor{gray!20}{70} &  \cellcolor{gray!20}{\textbf{48.2}} & \cellcolor{gray!20}{\textbf{70.8}} & \cellcolor{gray!20}{\textbf{53.0}} & \cellcolor{gray!20}{\textbf{43.7}} & \cellcolor{gray!20}{\textbf{66.7}} &  \cellcolor{gray!20}{\textbf{46.9}}\\
\Xhline{1.0pt}
\end{tabular}
}
\vspace{-3mm}
\caption{\textbf{Results of object detection, instance segmentation on COCO val2017}. AP$^\textrm{b}$ and AP$^\textrm{m}$ indicates box mAP and mask mAP, respectively. We measure FLOPs at $800\times1280$ resolution.} 
\label{tab:supp_coco}
\vspace{-2mm}
\end{table}

\begin{table}[t]
\setlength\tabcolsep{1.8pt}
\centering
\scalebox{0.90}{
\begin{tabular}[t]{L{3.85cm}|C{1.7cm}C{1.6cm}C{1.5cm}}
\Xhline{1.0pt}
\multirow{2}*{method} & \multicolumn{3}{c}{Semantic FPN 80K} \\
&\#param (M) & FLOPs (G) & mIoU (\%) \\
\Xhline{0.7pt}
Res101~\cite{he2015delving} & 48 & 260 & 38.8 \\
PVT-M~\cite{pvt} & 48 & 219 & 41.6 \\
PVT-L~\cite{pvt} & 65 & 283 & 42.1 \\
Swin-S~\cite{swin} & 53 & 274 & 45.2 \\
Twins-B~\cite{twins} & 60 & 261 & 45.3 \\
TwinsP-L~\cite{twins} & 65 & 283 & 46.4 \\
UniFormer-S$_{h32}$~\cite{uniformer}  & 25 & 199 & 46.2  \\
UniFormer-S~\cite{uniformer}  & 25 & 247 & 46.6  \\
\cellcolor{gray!20}{StructViT-S-4-1$_{h32}$} & \cellcolor{gray!20}{26} & \cellcolor{gray!20}{271} & \cellcolor{gray!20}{\textbf{46.9}} \\
\Xhline{0.7pt}
X101-32x4d~\cite{resnext} & 86 & - & 40.2 \\
Swin-B~\cite{swin} & 91 & 422 & 46.0 \\
Twins-L~\cite{twins} & 104 & 404 & 46.7 \\
UniFormer-B$_{h32}$~\cite{uniformer}  & 54 & 350 & 47.7  \\
UniFormer-B~\cite{uniformer}  & 54 & 471 & 48.0  \\
\cellcolor{gray!20}{StructViT-B-4-1$_{h32}$} & \cellcolor{gray!20}{54} & \cellcolor{gray!20}{529} & \cellcolor{gray!20}{\textbf{48.5}} \\
\Xhline{1.0pt}
\end{tabular}
}
\vspace{-3mm}
\caption{\textbf{Results of semantic segmentation on ADE20K.} We measure FLOPs using $512 \times 2048$ resolution images.}
\label{tab:supp_ade20k}
\vspace{-5mm}
\end{table}

$c^{\mathrm{struct}}_{i,j}$ indicates the final attention score multiplied to the value $\vv_j$ to generate the output.
 The examples in Fig.~\ref{fig:supp_image_visualization:f} show that StructSA contextualizes the entire features in a structure-aware manner considering objects' layouts or shapes; for instance, StructSA aggregates global contexts distinguishing different parts of an orange (Fig.~\ref{fig:supp_image_visualization:f},  2nd row) or an ostrich (Fig.~\ref{fig:supp_image_visualization:f}, 3rd row).
The qualitative analysis demonstrates that StructSA outperforms ConvSA in leveraging correlation structures for visual representation learning.
This suggests that StructSA may be particularly useful for computer vision tasks that require an understanding of relational structures and layouts of visual elements. 

\end{document}